\documentclass[9.5pt,journal,compsoc]{IEEEtran} 
\usepackage[pdftex]{graphicx}
\usepackage{booktabs, makecell, tabularx}
\usepackage{rotating}
\usepackage[export]{adjustbox}
\usepackage{gensymb}
\usepackage{ragged2e}
\usepackage{paralist}
\usepackage{multirow}
\usepackage{color}
\usepackage[pagebackref=true,breaklinks=true,letterpaper=true,colorlinks,bookmarks=false]{hyperref}

\definecolor{todocolor}{RGB}{0,174,247}

\newcommand{\whu}{FE-Blurframe} 
\ifCLASSOPTIONcompsoc
  \usepackage[nocompress]{cite}
\else
  \usepackage{cite}
\fi

\usepackage{xspace}
\usepackage{pgfplots}
\usepackage{xcolor}
\usepackage{tikz}
\usepackage{threeparttable}
\usepackage{stfloats}
\usepackage{amsmath}
\usepackage{amssymb}

\usepackage{array}

\ifCLASSOPTIONcompsoc
  \usepackage[caption=false,font=footnotesize,labelfont=sf,textfont=sf]{subfig}
\else
  \usepackage[caption=false,font=footnotesize]{subfig}
\fi

\ifCLASSOPTIONcaptionsoff
  \usepackage[nomarkers]{endfloat}
 \let\MYoriglatexcaption\caption
 \renewcommand{\caption}[2][\relax]{\MYoriglatexcaption[#2]{#2}}
\fi

\usepackage{url}
\usepackage{etex}

\renewcommand{\baselinestretch}{.985}

\begin{document}
\bstctlcite{IEEEexample:BSTcontrol}

\title{Detecting Line Segments \\ in Motion-blurred Images with Events}

\author{Huai Yu, 
        Hao Li, 
        Wen Yang, 
        Lei~Yu,~
        and~Gui-Song Xia

\IEEEcompsocitemizethanks{
\IEEEcompsocthanksitem H. Yu, H. Li, W. Yang, L. Yu are with the School of Electronic Information, Wuhan University, Wuhan 430072, China. 
E-mail: \{yuhuai, lihao2015, yangwen, ly.wd\}@whu.edu.cn.
\IEEEcompsocthanksitem G. S. Xia is with the School of Computer Science, Wuhan University, Wuhan 430072, China.
E-mail:guisong.xia@whu.edu.cn.
\IEEEcompsocthanksitem H. Li is also with Baidu Inc, Beijing, China.
\IEEEcompsocthanksitem H. Yu and H. Li contribute equally to this work.
}
}

%



\newcommand{\ie}{\textit{i.e.}\xspace}
\newcommand{\eg}{\textit{e.g.}\xspace}
\newcommand{\etal}{\textit{et al.}\xspace}
\newcommand{\wrt}{\textit{w.r.t.}\xspace}
\newcommand{\etc}{\textit{etc.}\xspace}
\newcommand{\resp}{\textit{resp.}\xspace}

\newcommand{\fixedvskip}{-3mm}
\newcommand{\fixedvskiptab}{-3mm}

\newcommand{\equspace}{6pt}
\newcommand{\revision}[1]{{\color{blue}{#1}}}
\newcommand{\smallrevision}[1]{{\color{black}{#1}}}



\definecolor{todocolor}{RGB}{0,174,247}

\IEEEtitleabstractindextext{%
\begin{abstract}
    \justifying
Making line segment detectors more reliable under motion blurs is one of the most important challenges for practical applications, such as visual SLAM and 3D reconstruction. Existing line segment detection methods face severe performance degradation for accurately detecting and locating line segments when motion blur occurs. While event data shows strong complementary characteristics to images for minimal blur and edge awareness at high-temporal resolution, potentially beneficial for reliable line segment recognition. 
To robustly detect line segments over motion blurs, we propose to leverage the complementary information of images and events. To achieve this, we first design a general frame-event feature fusion network to extract and fuse the detailed image textures and low-latency event edges, which consists of a channel-attention-based shallow fusion module and a self-attention-based dual hourglass module. We then utilize the state-of-the-art wireframe parsing networks to detect line segments on the fused feature map. Moreover, due to the lack of line segment detection dataset with pairwise motion-blurred images and events, we contribute two datasets, \ie, synthetic FE-Wireframe and realistic \whu, for network training and evaluation. Extensive experiments on both datasets demonstrate the effectiveness of the proposed method. When tested on the real dataset, our method achieves $63.3\%$ mean structural average precision (msAP) with the model pre-trained on the FE-Wireframe and fine-tuned on the \whu, improved by $32.6$ and $11.3$ points compared with models trained on synthetic only and real only, respectively. The codes, datasets, and trained models are released at \url{https://levenberg.github.io/FE-LSD}.
\end{abstract}

\begin{IEEEkeywords}
Line segment detection; Motion blur;  Frame-event fusion; Attention.
\end{IEEEkeywords}}

\maketitle

\IEEEdisplaynontitleabstractindextext

%
\IEEEpeerreviewmaketitle


\section{Introduction}
\label{sec:introduction}

\begin{figure}[!t]
	\centering
	\includegraphics[width=0.95\linewidth]{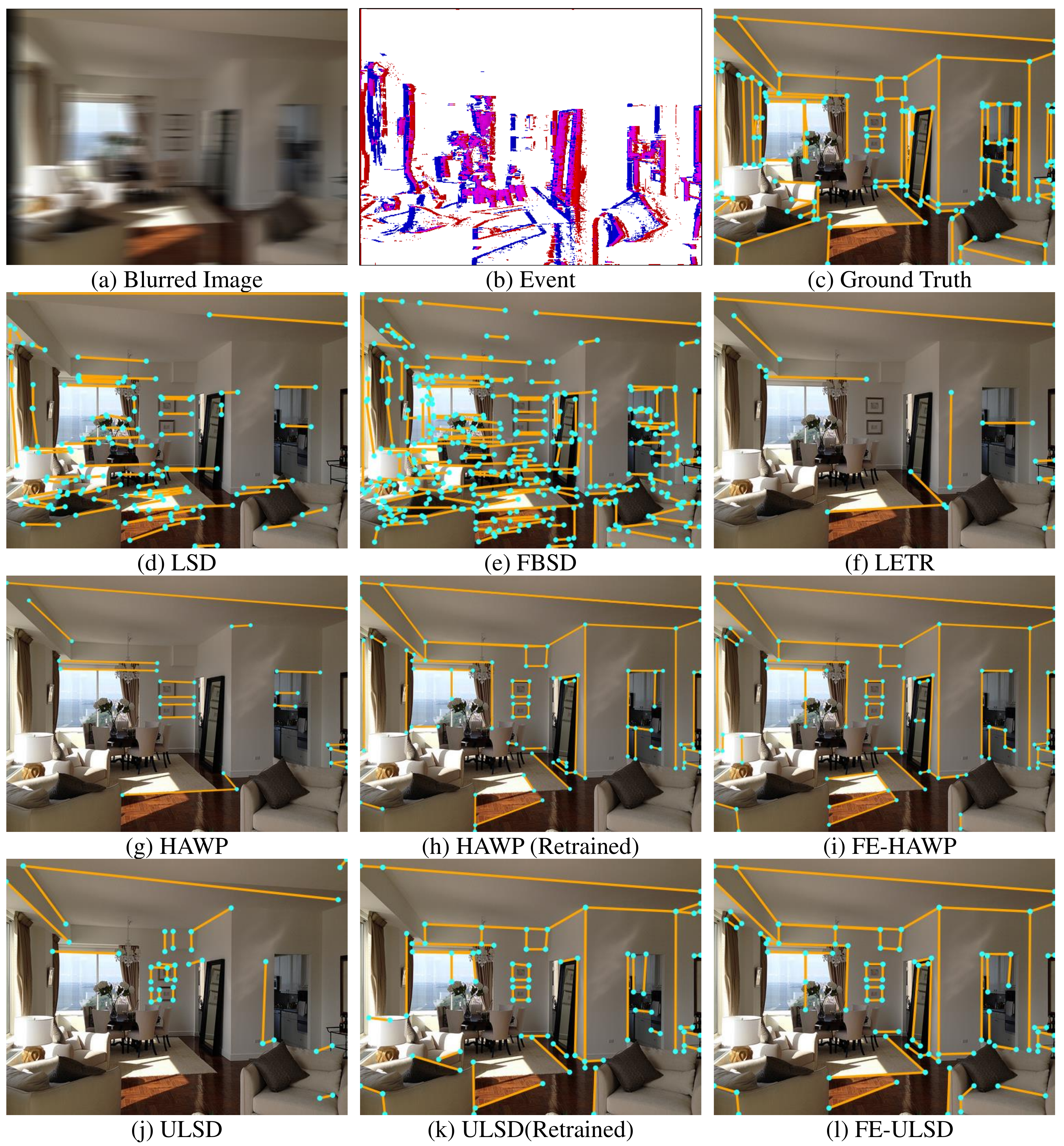} 
	\caption{Illustrative examples of different methods. Line segments are drawn on clear images at the end of camera exposure for better visualization. Conventional (d)LSD \cite{von2010lsd} and (e)FBSD \cite{even2019thick} have many false detections while the original (f)LETR \cite{xu_line_2021}, (g)HAWP \cite{xue2020holistically} and (j)ULSD \cite{li2021ulsd}  have many missing alarms on the (a)motion-blurred image. The retrained (h)HAWP and (k)ULSD on the concatenation of (a)images and (b)events detect more line segments. The proposed (i)FE-HAWP and (l)FE-ULSD have the best performance.}
	\label{fig:teaser}
\end{figure}

\IEEEPARstart{L}{ine} segment detection in images with the emergence of motion blurs often suffers from severe performance degradation due to the motion ambiguities and texture erasures~\cite{jin2018learning}, which brings a serious burden to geometric structure perception especially when encountering high-speed motions of imaging platforms, \eg, robotics~\cite{Munir2022LDNet}, visual SLAM \cite{gomez2019pl}, and autonomous driving \cite{wang2022Multitask}. In contrast to those in images with clear and sharp edges, geometric structures in images with motion blurs are often blurred out, and it poses intense challenges to accurately detect and localize line segments in the images, see Fig. \ref{fig:teaser} (a) for an instance. 


%
Although promising progress has been reported in the past decades on line segment detection, see \eg,~\cite{von2010lsd,xue2020holistically,li2021ulsd, xu2021line}, few of them consider the scenarios with blurry textures and diverse motions, thus often losing their 
effectiveness when dealing with motion-blurred images. As illustrated in Fig. \ref{fig:teaser}, due to the overlapping of scene information at different times in blurriness, line segments no longer appear as distinctive elongated edges, while as blurred bands mixed with the background. To handle this issue, an intuitive solution is to deblur the image first and then detect line segments \cite{xu20112d}. The detection performance highly depends on the quality of motion deblurring, which itself is ill-posed due to motion ambiguities. Moreover, motion deblurring methods are usually developed for improving the image quality but not designed for a better structure perception \cite{jin2018learning,purohit2019bringing}. 
On the other hand, if directly performing the detection in motion-blurred images, we can only determine the distribution or the blurred band of each line segment \cite{debled2006optimal}. To measure the distribution, a thickness parameter is introduced to describe the line width \cite{buzer2007simple, even2019thick}. Nevertheless, it can neither model diverse blurred lines (\eg, rotated around a point on lines) nor geometrically locate the position of a line segment. Even if utilizing learning-based models \cite{zhou_end2019, xu2021line}, training with line annotations is still unable to model the blur distribution and determine the line position. Therefore, it is far from satisfactory to detect only on blurred images, regardless of whether deblurring, motion modeling, or data-driven approaches are used.

Recently, the newly developed event cameras have attracted increasing attention, outputting pixel-level brightness changes instead of standard intensity frames \cite{tulyakov2021time}. Event data is asynchronously generated at intensity edges with extremely high temporal resolution (up to $\mu$s), thus it is edge-aware and free of motion blurs \cite{wang2022Multitask}. Motivated by this characteristic, we propose to introduce event data into the task of line segment detection in motion-blurred images. On one hand, texture and structure layout can be maintained by image frames, which also help to suppress the effect of event noise. On the other hand, event data can recognize the distinctive edges in blurred images and address diverse camera motions. 
Consequently, event cameras show strong complementary characteristics to RGB cameras for their minimal blur and edge awareness at high-temporal resolution.
Therefore, if events can be used in the data-driven training process, we will have great potential to improve the robustness of line detection and localization to motion blur.  However, it remains challenging for the effective fusion of the two domain data to detect line segments. Several key problems need to be investigated: how to avoid the influence of event noises on line endpoint prediction, how to make the fused feature map be cognitive to line edges with either fast or slow camera motions, and how to solve the problem of lacking line segment detection datasets with both visual and event data.



In this paper, we propose a robust line segment detection method in motion-blurred images with events, \ie, Frame-Event fused Line Segment Detection (FE-LSD). As the frontend, a novel feature fusion backbone is designed to extract and fuse information from images and events, which consists of a channel-attention-based shallow module and a self-attention-based stacked dual hourglass module. The shallow fusion module is designed to suppress event noise and enhance image edge features. The stacked dual hourglass module is developed to conduct multi-scale fusion and obtain the fused feature map.
As the backend, the state-of-the-art line segment detectors, \eg, HAWP and ULSD, are used to detect line segments on the fused feature map, namely FE-HAWP and FE-ULSD respectively. A glance at the performance of different methods is shown in Fig. \ref{fig:teaser} (i) and (l), where the proposed FE-HAWP and FE-ULSD have the best performance. As far as we know, this is the first work of robust line segment detection in motion-blurred images with events. Our contributions can be summarized as follows:

\begin{itemize}
    \item Events are introduced to assist motion-blurred images for line segment detection, which can robustly address the performance degradation caused by motion blurs. The idea of fusing images and events fully exploits their complementary properties of low-latency event edges and detailed image textures, thus effectively improving the accuracy and robustness of line detection over diverse camera motions.
    \item A general frame-event feature fusion network is designed to extract and fuse the information from images and events. The concatenation of shallow fusion and multi-scale decoder fusion fully explores the channel-attention and self-attention mechanism, thus enhancing the feature extraction of events and frames.
    \item Two frame-event line segment detection datasets, \ie, synthetic FE-Wireframe and real-world \whu, are constructed for line segment detection in motion-blurred images. Sufficient qualitative and quantitative comparisons demonstrate the effectiveness and robustness of the proposed method.
\end{itemize}


\section{Related Work}
\label{sec::relatedwork}
Line segment detection has been a focused fundamental problem in computer vision over decades, especially with the revolution of deep learning. In this section, we will review the image-based and event-based line segment detection methods. Image-based methods recall the track from traditional edge-based clustering to the latest learning-based algorithms. While for the event-based methods, we mainly analyze the recent adaptions of frame-based methods to events, such as Hough transform \cite{seifozzakerini2016event} and LSD \cite{brandli2016elised}.

\subsection{Image-based line segment detection}
Traditional line detection methods mostly rely on edge detection and aggregation of linear geometric properties, such as Hough transform \cite{xu_accurate_2015,almazan_mcmlsd_2017}, LSD \cite{von2010lsd}, EDLines\cite{akinlar_edlines_2011}, and  linelet\cite{cho_novel_2018}. These methods often follow an unsupervised manner on the CPU and thus can be efficiently deployed on embedding systems. However, their performance highly depends on the edge quality and is sensitive to motion blurs. In recent years, learning-based techniques have shown great progress to detect line segments. The first learning-based wireframe parsing algorithm DWP \cite{huang2018learning} fuses the predicted line segments and junctions with two CNNs to obtain the line segments. Subsequently, AFM \cite{xue2021afm} is proposed to transform this task into an attraction field regression problem using the line attraction field representation. Those two methods are the first to introduce deep learning to improve the performance of line segment detection. However, they both rely on heuristic post-processing and thus are not end-to-end. To address this issue, a series of end-to-end line segment detection methods have been proposed \cite{zhou_end2019, xue2020holistically, zhang2019ppgnet,meng020lgnn}. The first one, L-CNN\cite{zhou_end2019} utilizes a proposal sampling module to conduct the fusion of lines and junctions in the network, eliminating the non-differentiable heuristic post-processing and ensuring it is fully differentiable. After that, the HAWP \cite{xue2020holistically} combines the attraction field map \cite{xue2021afm} with the L-CNN network, which effectively improves the quality of line segment proposals. Subsequently, a trainable Hough transform module is added to the existing line segment detection network \cite{lin2020deep} to enhance the linear geometric constraints. Meanwhile, considering that line segments formulate graphs, PPGNet\cite{zhang2019ppgnet} and LGNN\cite{meng020lgnn} introduce Graph Neural Network (GNN) to fuse the line segments and junctions for optimization, which eventually yields more accurate line detection results. Recently, Transformer was introduced to the line segment detection network. By combining it with CNN, LETR \cite{xu2021line} achieves state-of-the-art performance in an end-to-end manner. However, its main drawback is that model training is time-consuming, and inference is relatively slow. Recently, self-supervised \cite{pautrat2021sold2} and semi-supervised \cite{kong2022hole} methods have also been proposed to overcome the problem of limited training samples. 

These learning-based line segment detection methods have demonstrated state-of-the-art performance. However, they are trained and tested on clear images without motion blurs. For real-world applications such as mobile robotics and autonomous driving, the fast relative motion between the camera and objects will inevitably generate motion blurs in RGB images. The blurriness can significantly compromise or even invalidate the performance of existing methods.

 There are few existing works for line segment detection in motion-blurred images. Debled-Rennesson first presents the concept of blurred line segments in images and gives its mathematical definition \cite{debled-rennesson2003segmentation} by introducing a thickness parameter to describe the line width. 
Then this blurred line representation model is further extended to gray-scale images, and a semi-automatic line segment selection tool is implemented \cite{kerautret2009blurred}. Later, FBSD \cite{even2019thick} is proposed based on the adaptive directional scans and the control of assigned thickness to achieve automatic blurred line segment detection.
 Compared with the traditional LSD\cite{von2010lsd} and EDLines\cite{akinlar_edlines_2011}, FBSD effectively improves the line location accuracy and recall when handling blurs. However, the blurred line definition according to Debled-Rennesson can only determine the width bands of line segments. Thus it is hard to give an accurate line location with large motion blurs.

\subsection{Event-based line segment detection}
Over the past few years, several event-based line segment detection methods \cite{reverter2019event,tschopp2021hough2map,Munir2022LDNet} have been proposed using the newly developed event camera, which does not suffer motion blur due to its asynchronous characteristics and high temporal resolution \cite{gallego2022event}. The Hough transform is the first method \cite{seifozzakerini2016event} for event-based line segment detection and has recently been used for high-speed train localization and mapping \cite{tschopp2021hough2map}, demonstrating the effectiveness in handling high-speed motions. After that, the ELiSeD \cite{brandli2016elised} is proposed by adapting LSD \cite{von2010lsd} to the event camera. Based on the assumption of a constant velocity model during a short-time motion period, the fast detection and persistent tracking of translating line segments are achieved by plane detection in event data stream \cite{everding2018low}. However, it cannot handle rotating and drastically moving line segments. To improve the robustness over motion speed, an event-based line segment detection method using iteratively weighted least-squares fitting was proposed \cite{reverter2019event}. Recently, DVS event cameras are used for lane detection using basic CNN and transformer networks\cite{Munir2022LDNet, cheng2019det}, which give some insights for learning-based line detection. In \cite{ dietsche2021powerline}, an online event-based powerline tracker is proposed by using hibernation to cope with the line direction changes. Most of these methods are validated only on simple data, thus lacking quantitative evaluation and comparison. Additionally, there is no publicly available dataset as the testing benchmark, and the line segment detection performance in real environments is yet to be verified.  

Although event cameras do not suffer motion blurs, the line segment detection performance on pure event stream is also unsatisfactory because of the lack of texture information and instability to slow motions or static conditions. Detailed texture can improve the performance in weakly textured areas and maintain structural integrity. Slow motions or static event cameras cannot capture informative events. In contrast, RGB cameras can clearly capture detailed edges and do not have motion blur when moving slowly and being stationary. Therefore, we propose to fully utilize the strengths of both cameras to achieve line segment detection in diverse motion conditions by fusing visual images and event streams. However, this task is challenging, such as the lack of frame-event fusion strategy for geometric structure perception, and the lack of line segment detection dataset with visual images and event stream data.

\section{Methodology}
\label{sec::methodology}

\subsection{Problem statement}
We first introduce the concept of blurred line segments proposed by Debled-Rennesson \cite{debled-rennesson2003segmentation}. A discretized straight line is defined as:
\begin{equation}
	\label{eq:blurredlinedef}
	l = \{(x,y) | c \leq ax - by < c+v\}, 
\end{equation}
where $(x,y) \in \mathbb{Z}^2$ is the pixel coordinate, $\{a, b, c, v\} \in \mathbb{Z}^4$ are the four parameters to determine the line segment, in which $v$ describes line width. An additional thickness parameter $\mu$ is defined as the minimum of the horizontal and vertical distances between the lines $ax - by$=$c$ and $ax - by$ = $c + v$, i.e., $\mu$ = $\frac{v}{max(|a|,|b|)}$.


\begin{figure}[htb]
	\begin{minipage}[t]{0.32\linewidth}
		\centering
		\includegraphics[width=\columnwidth]{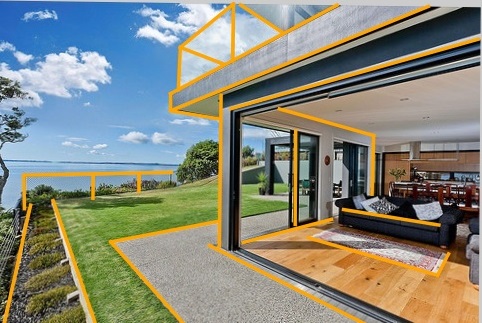} 
		\centerline{\footnotesize{(a) Start exposure}}
		\label{fig:startexpo}
	\end{minipage}
	\begin{minipage}[t]{0.32\linewidth}
		\centering
		\includegraphics[width=\columnwidth]{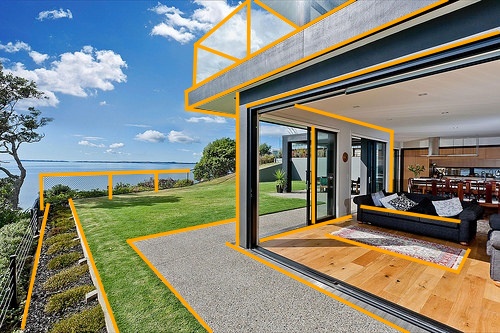} 
		\centerline{\footnotesize{(b) End exposure}}
		\label{fig:endexpo}
	\end{minipage}
	\begin{minipage}[t]{0.32\linewidth} 
		\centering
		\includegraphics[width=\columnwidth]{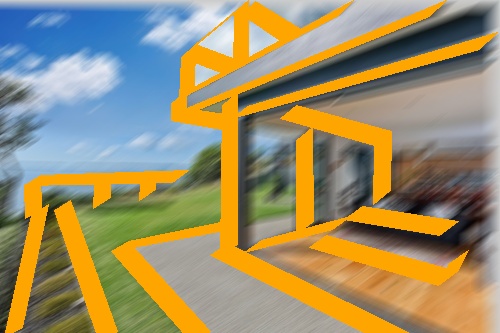} 
		\centerline{\footnotesize{(c) During exposure}}
		\label{fig:blurredlines}
	\end{minipage}
	\caption{Blurred line segments in motion-blurred images. From left to right are the line segments at the beginning, end, and within the camera exposure time. }
	\label{fig:blurlinedef}
\end{figure}
In fact, the blurry distribution of line segments is generally complex in motion-blurred images, as shown in Fig. \ref{fig:blurlinedef}. A simple thickness parameter cannot precisely model the line distributions. 
In this paper, line segments in images are defined as a function of time $l(t), 0 \leq t \leq T$, which are detected on the clear image captured at time $t$. $T$ is the camera exposure time. The imaging process for motion-blurred images can be treated as the average of all clear images $I(t)$ during $T$, i.e., $I_B$=$\frac{1}{T}\int_{t=0}^{T} I(t) dt$. Since the temporal resolution is lost by averaging over time, it is difficult to detect line segments by only relying on the motion-blurred image. Therefore, we introduce the high-temporal resolution events $\mathcal{E}$=$\{(x_i,y_i,p_i,t_i)\}_{t_i \in T}$ into this task. Given the motion-blurred image $I_B$ and the temporally aligned events $\mathcal{E}$ during $T$, the line segment detection with the frame and event data can be defined as:
\begin{equation}
	\label{eq:linedef2}
	\mathcal{L}(T) = \mathrm{FE}\text{-}\mathrm{LSD}(I_B, \mathcal{E}),
\end{equation}
where $\mathcal{L}(T)$ denotes the set of line segments at time $T$ and $\mathrm{FE}$-$\mathrm{LSD}$ denotes the line segment detection function.



\subsection{Event representation}
Unlike RGB cameras streaming images at a certain frequency, event cameras do not output synchronous frames but event points asynchronously. 
For each pixel $\boldsymbol{u}$=$\left[x,y\right]^T$, if the captured luminance $L$ changes beyond a contrast threshold $C$ at time $t$, an event point $e$=$(x,y,t,p)$ is triggered. The increase and decrease of luminance will result in positive ($p$=$+1$) and negative ($p$=$-1$) polarities, respectively.

\begin{equation}
	\label{eq:eventgen}
	 \Delta L = p \left(L(x,y,t) - L(x,y,t-\Delta t)\right) \geq C,
\end{equation}
where $t-\Delta t$ is the time of the last event point generated at that pixel.

\begin{figure}[htbp]
	\begin{minipage}[t]{0.49\linewidth}
		\centering
		\includegraphics[width=0.99\columnwidth]{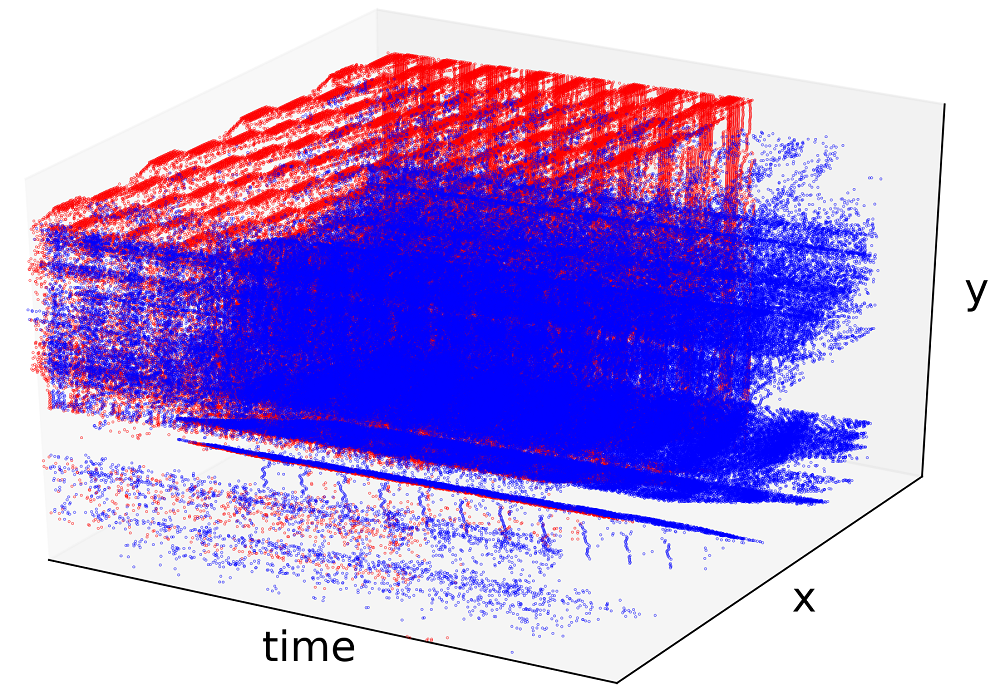} 
		\centerline{\footnotesize{(a) Raw event stream}}
	\end{minipage}
	\begin{minipage}[t]{0.49\linewidth}
		\centering
		\includegraphics[width=0.9\columnwidth]{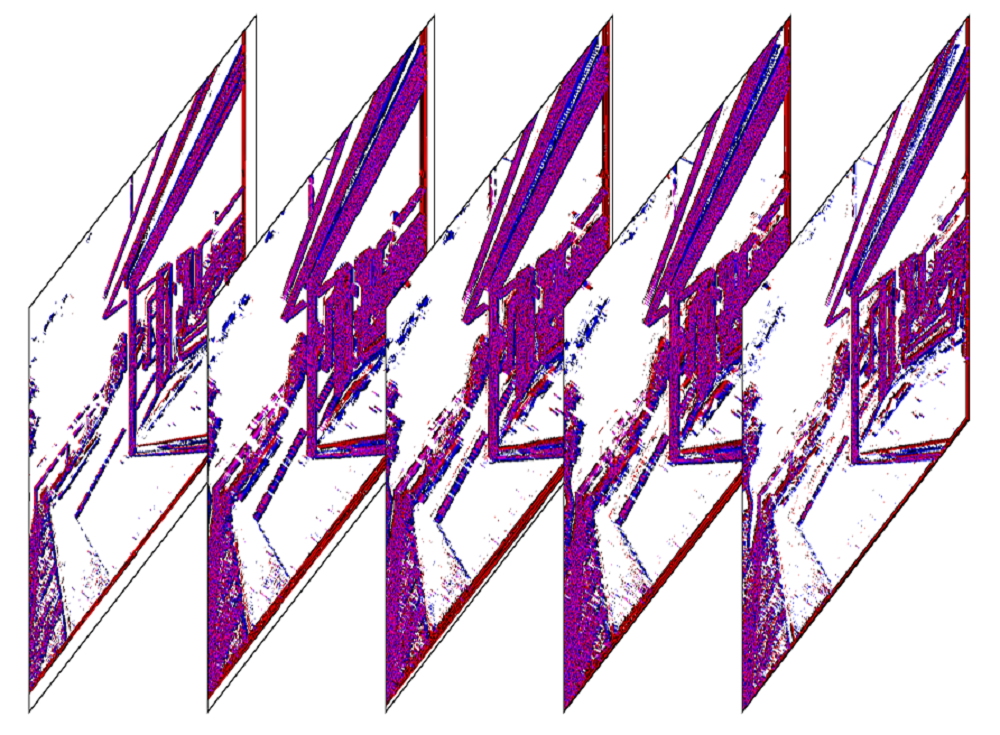} 
		\centerline{\footnotesize{(b) EST representation}}
	\end{minipage}
	\caption{Event data representation using the EST (Left: raw event stream, red-positive events, blue-negative events; Right: EST representation with $B=5$).}
	\label{fig:evframerec}
\end{figure}

\begin{figure*}[!b]
	\centering
	\includegraphics[width=1.0\linewidth]{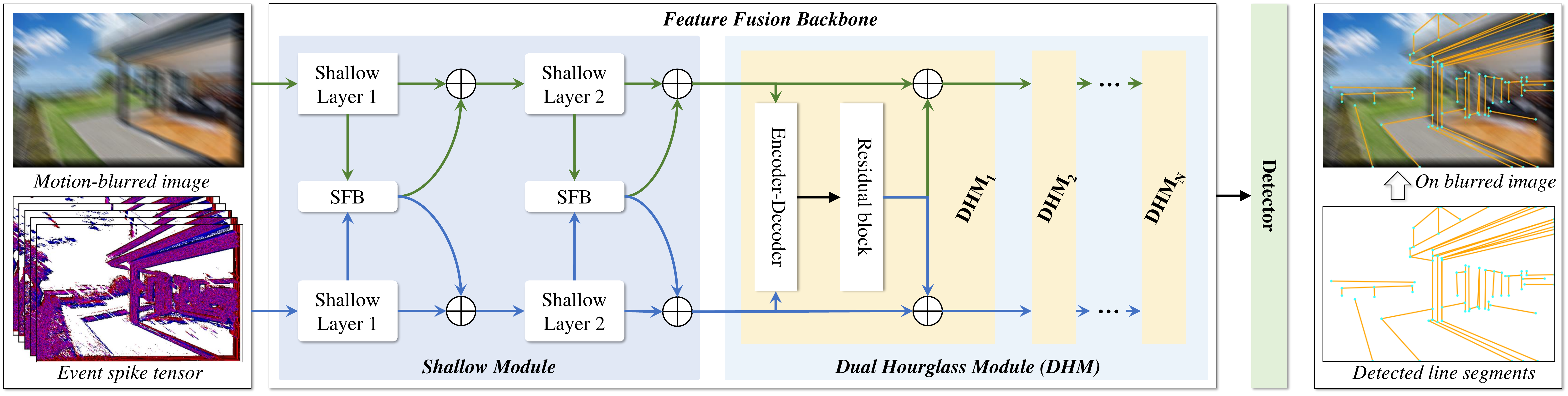} 
	\caption{FE-LSD network structure. It mainly consists of two modules: feature fusion backbone and back-end line detector, in which the feature fusion backbone has one shallow module and several dual hourglass modules.}
	\label{fig:FE-LSD-net}
\end{figure*}
Event streams have asynchronous and sparse characteristics. To follow the CNN architecture, the asynchronous event streams are often transformed into fixed-size tensors. The most commonly used event representations including Event Counting (EC) \cite{maqueda2018event}, Surface of Active Event (SAE) \cite{benosman2013event}, voxel grid \cite{zhu_unsupervised_2019} and Event Spike Tensor (EST) \cite{gehrig_end--end_2019}. Among these the EST is a four-dimensional grid of $H$$\times $$W$$\times$$B$$\times$$2$, where the time dimension is uniformly divided into $B$ bins. Each bin further encodes the polarity as the positive and negative dimensions. Therefore, it retains the most temporal and polarized information. The EST is calculated by: 
\begin{equation}
	\label{eq:EST}
	\mathbf{EST}_{\pm}(x, y, b) = \sum_{e_{k} \in \mathcal{E}_{\pm}} \delta\left(x-x_{k}, y-y_{k}\right) \max \{0,1-|b-t_{k}^{*}|\},
\end{equation}
where $b$$\in$$\{0, 1, \cdots, B-1\}$, $t_{k}^{*} \triangleq \frac{B-1}{T}\left(t_{k}-t_{0}\right)$, $t_0$ is the earliest timestamp of the event stream during time $T$, $e_{k} \in \mathcal{E}_{\pm}$ means the positive and negative EST are computed respectively with the polarities of events. Fig. \ref{fig:evframerec} shows the event reconstruction based on the EST representation. The two spatial-temporal lattices corresponding to the positive and negative polarity of EST is further stacked into a $H$$\times$$W$$\times$$2B$ tensor to fit the common requirement of CNN.


\subsection{Network structure}
Given the motion-blurred image and the aligned EST data, an effective fusion strategy is designed to extract the complementary information for line segment detection. We constructively fuse the shallow and deep features from images and events, and then use two state-of-the-art line segment detectors, \ie, HAWP \cite{xue2020holistically} and ULSD \cite{li2021ulsd} to get the final detection results (Fig. \ref{fig:FE-LSD-net}). The feature fusion backbone network includes two kinds of modules: (\romannumeral1) Shallow module, which is designed to extract shallow features and suppress the event noises, and (\romannumeral2) Dual hourglass module, which follows an encoder-decoder structure to conduct a multi-scale feature fusion for the feature map generation.


\subsubsection{Shallow module}
With the resized $H$$\times$$W$$\times$$3$ RGB image and the spatially-aligned $H$$\times$$W$$\times$$2B$ EST data, the shallow module first separately down-samples the RGB and EST using the Shallow Layer 1 (including 7$\times$7 convolution with step size 2, BatchNorm, and ReLU), which can obtain the preliminary RGB and EST features with the same number of channels. Then the image and event features are fused by the Shallow Fusion Block (SFB), and the fused feature is further added with the original RGB and EST features, respectively. Next, the Shallow Layer 2 (three residual blocks \cite{he_deep_2016} and one maximum pooling module after the first residual block) is used to down-sample and extract features. The refined features are inputted into the second SFB module. Finally, the fused feature is added with the outputs of Shallow Layer 2 to obtain the same dimension image feature $X_F^{(0)}$ and the event feature $X_E^{(0)}$, which will be fed to the dual hourglass module subsequently.

\begin{figure}[htbp]
	\centering
	\includegraphics[width=0.9\linewidth]{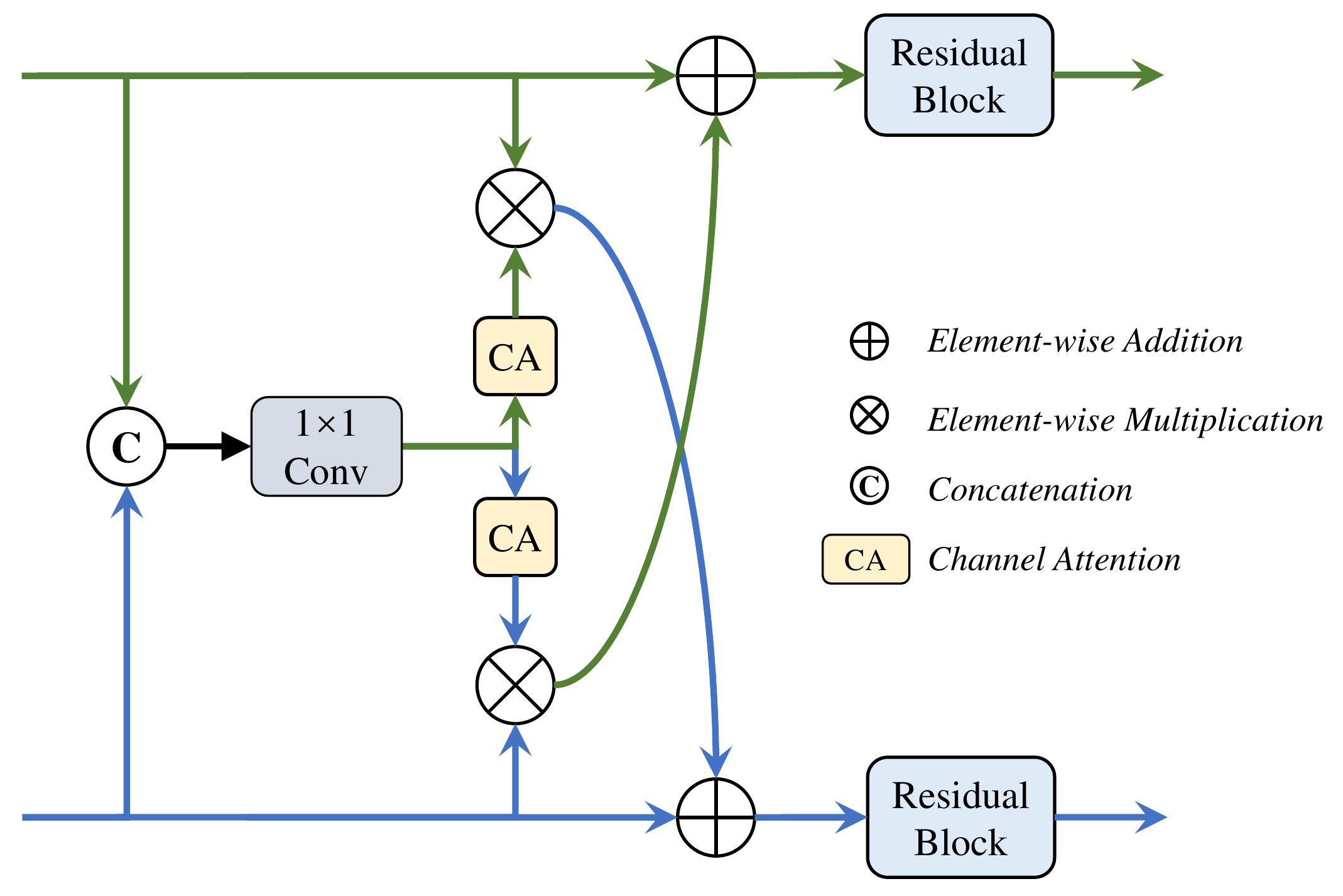} 
	\caption{Shallow Fusion Block (SFB).}
	\label{fig:SFB}
\end{figure}
The SFB is the core part of the shallow module, and its network structure is shown in Fig.~\ref{fig:SFB}. For the input image feature $X_F$ and the EST feature $X_E$, the SFB first concatenates them together by channels and retains the number of channels with a 1$\times$1 convolution. Then the attention $Attn_F$ and $Attn_E$ of the two-channel features are computed separately using the channel-attention (CA) blocks. The attention is further multiplied with the original features and then added with the original features of the other modality to achieve attention-weighted feature fusion. Finally, the fused features are refined using a residual block to obtain the shallow image feature $X^o_F$ and event feature $X^o_E$. The whole processing process of the SFB can be described by 
\begin{equation}
	\label{eq:shallowmodule}
	\begin{aligned}
		X &= \mathrm{Conv_{1 \times 1}}\left(\mathrm{Concat}(X_F,X_E)\right) \\
		Attn_F &= \mathrm{CA}_F(X) \\
		Attn_E &= \mathrm{CA}_E(X) \\
		X^o_F &= \mathrm{Res}_F(X_F + X_E \odot Attn_E) \\
		X^o_E &= \mathrm{Res}_E(X_E + X_F \odot Attn_F),
	\end{aligned}
\end{equation}
where $\mathrm{Conv_{1\times 1}}$ denotes 1$\times$1 convolution, $\mathrm{Concat}$ denotes the concatenation by channel, $\mathrm{CA}$ denotes the channel-attention module, $\mathrm{Res}$ denotes the residual module and $\odot$ denotes the element-wise multiplication.

\subsubsection{Dual hourglass module}
As in HAWP \cite{xue2020holistically} and ULSD \cite{li2021ulsd}, an encoder-decoder network, i.e., stacked hourglass network \cite{leibe_stacked_2016}, is used to obtain the feature map for line segment detection. With the two-branch shallow features, we design the stacked dual hourglass module to further fuse and extract features, as shown in Fig.~\ref{fig:FE-LSD-net}. The image and event features will be first fused into one feature with the encoder-decoder network, followed by a residual block. Then the fused feature is added with the input image and event features respectively to recover the two branch features for the next dual hourglass module. For the last dual hourglass module, the fused feature after the residual block is directly outputted as the final feature map for the subsequent line segment detector. The calculation of the stacked dual hourglass module is as:
\begin{equation}
	\label{eq:cascadhourglass}
	\begin{aligned}
		Y^{(i)} &= \mathrm{Res}(\mathrm{E \mbox{-} D}(X_F^{(i)},X_E^{(i)})) \\
		X_F^{(i + 1)} &= X_F^{(i)} + Y^{(i)} \\
		X_E^{(i + 1)} &= X_E^{(i)} + Y^{(i)},
	\end{aligned}
\end{equation}
where $\mathrm{E \mbox{-} D}$ denotes the encoder-decoder network, $X_F^{(i)}$ and $X_E^{(i)}$ denote the output image and event features of the $i$-th dual hourglass module, respectively ($X_F^{(0)}$ and $X_E^{(0)}$ are the outputs of the shallow module). $Y_i$ denotes the fused features of the $i$-th dual hourglass module output.
\begin{figure}[htbp]
	\centering
	\includegraphics[width=0.99\linewidth]{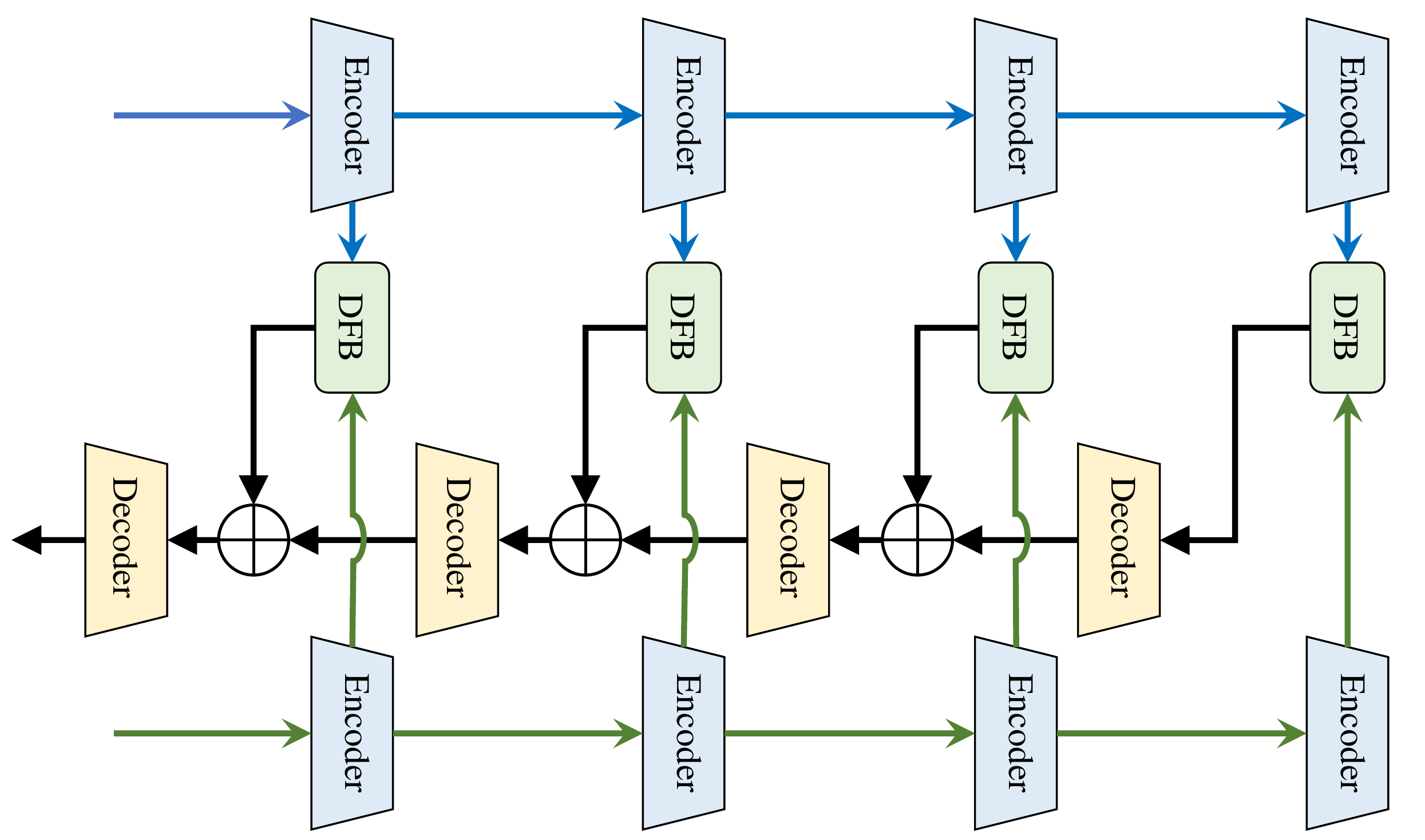} 
	\caption{The encoder-decoder network in the dual hourglass module.}
	\label{fig:encode-decoder}
\end{figure}

\begin{figure}[htbp]
	\centering
	\includegraphics[width=0.9\linewidth]{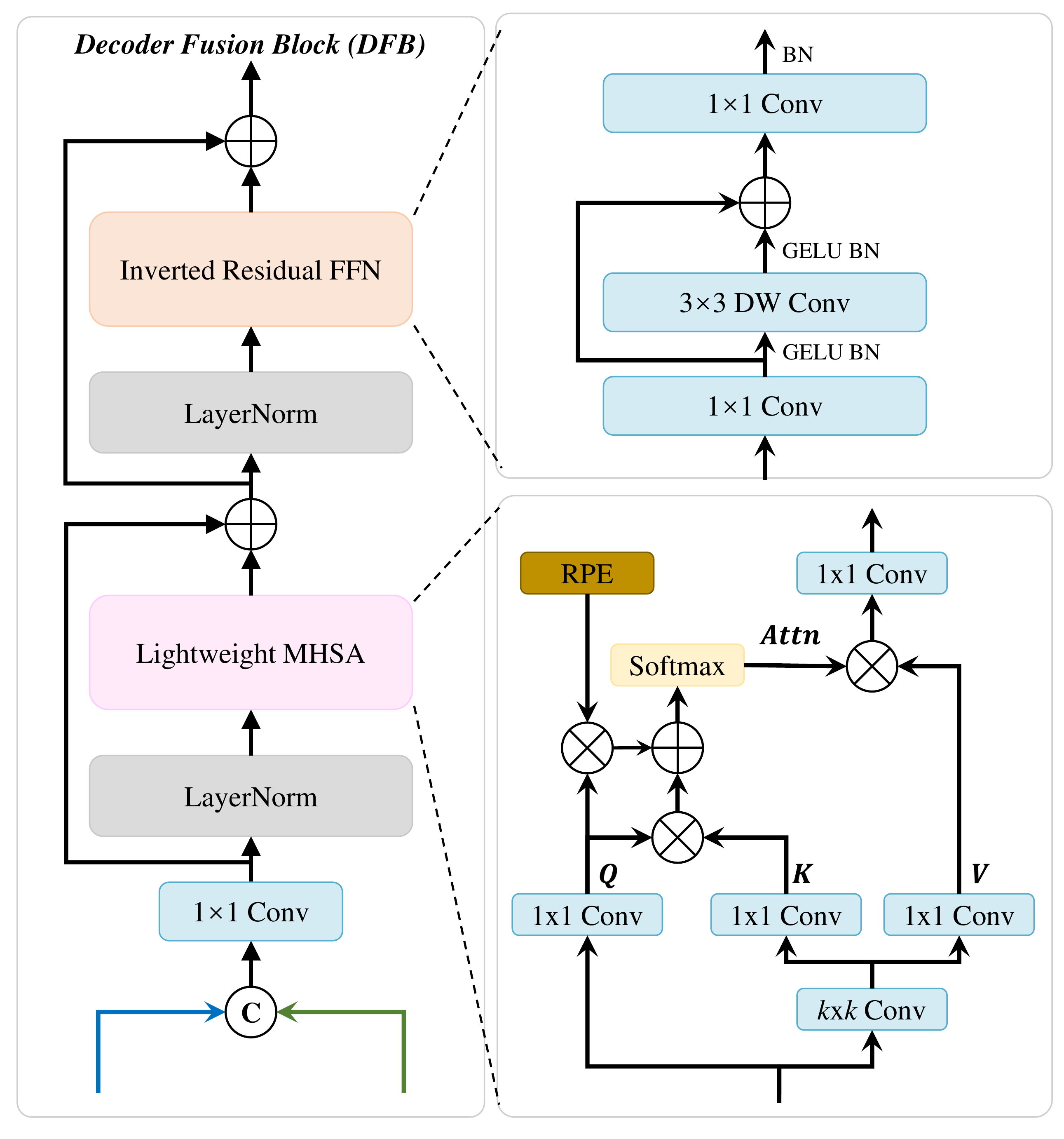} 
	\caption{The decoder fusion block in the encoder-decoder network.}
	\label{fig:DFB}
\end{figure}

The core of the dual hourglass module is the encoder-decoder network, as shown in Fig.~\ref{fig:encode-decoder}, which consists of a pairwise image feature encoder, event feature encoder, decoder fusion block (DFB), and decoder. Encoders and decoders are implemented by residual blocks. The feature map is down-sampled during feature encoding using a max-pooling layer with step size 2. During decoding, it is up-sampled with step size 2 to ensure the same feature size for the encoder, decoder, and DFB at the same level. The DFB (Fig.~\ref{fig:DFB}) consists of a 1$\times$1 convolution and a transformer module using 2 Layer Normalization (LN) modules, a lightweight Multi-head Self-attention (MHSA) module, and an Inverse Residual Feed-forward Network (IRFFN) \cite{guo_cmt_2021}. The DFB module first fuses the input image-event features to one feature by channel concatenation and 1$\times$1 convolution, and then uses the transformer \cite{vaswani2017attention} to further fuse and extract deep features. The transformer is used mainly for two reasons: (\romannumeral1) Compared with CNN, the transformer has a stronger capacity to capture global features, which is beneficial for the long-distributed target detection such as line segments; (\romannumeral2) For the fusion of image and event features, the transformer can provide self-attention as well as cross-attention for global cross-modal feature interaction and information fusion.

The structure of the Lightweight MHSA module is shown in the right-bottom of Fig. \ref{fig:DFB}. Compared with the original MHSA \cite{vaswani2017attention}, this lightweight version adds a $k \times k$ convolution (with step size $k$) to reduce the spatial size of $K$ and $V$, thus reducing the computational complexity and memory consumption. Additionally, a learnable relative position encoder is used to provide the relative position information between pixels. For the input feature $X \in \mathbb{R}^{h \times w \times c}$, the Lightweight MHSA module computes the output feature $Y \in \mathbb{R}^{h \times w \times c}$ as:
\begin{equation}
	\label{eq:MHSA}
	\begin{aligned}
		X^{\prime} &= \mathrm{Conv_{k \times k}}(X) \\
		Q &= \mathrm{Conv_{1 \times 1,Q}}(X) \\
		K &= \mathrm{Conv_{1 \times 1,K}}(X^{\prime}) \\
		V &= \mathrm{Conv_{1 \times 1,V}}(X^{\prime}) \\
		Attn &= \mathrm{Softmax}\left(\frac{Q \cdot {(K + \mathrm{RPE})}^T}{d_h}\right)  \\
		Y &= Attn \cdot V, 
	\end{aligned}
\end{equation}
where $\mathrm{RPE} \in \mathbb{R}^{\frac{h}{k} \times \frac{w}{k} \times c}$ is the relative position encoder, $N_h$ is the number of multiple heads, and $d_h$ is the dimension of each head.

After the features go through the Lightweight MHSA module, they are then fed to the IRFFN \cite{guo_cmt_2021}. It consists of 1$\times$1 convolutions and 3$\times$3 deepwise (DW) convolution, where the 3$\times$3 deep convolution can reduce the computational cost while extracting local features. The outputs of the first two convolutions are activated by GELU and BatchNorm.  The whole IRFFN computation process can be expressed as:
\begin{equation}
	\label{eq:IRFFN}
	\begin{aligned}
		X^{\prime} &= \mathrm{Conv}_{1 \times 1}(X) \\
		Y &= \mathrm{Conv}_{1 \times 1}(\mathrm{DWConv}(X^{\prime}) + X^{\prime}).
	\end{aligned}
\end{equation}

\subsubsection{Line Segment Detector}
After obtaining the fused feature map, we use state-of-the-art line segment detectors to conduct the line segment detection. The process is divided into two stages, i.e., the line proposal network and the line classification network. During the first stage, the junction proposal module and the line proposal module are used to generate junction and line proposals, respectively. Then the line and junction proposals are matched based on the connection relationship to generate the final line segment candidates. For the second stage, each line candidate is associated with a feature vector from the fused feature map by geometric matching. Then the line classification network is used to classify the line segments by ground truth binary supervision. Finally, candidates with confidence scores higher than the set threshold are selected as the final detection results.

By combining the feature fusion backbone with the line proposal and classification network, line segment detection in motion-blurred images can be achieved using synchronized visual images and EST data. The feature fusion backbone network can be easily applied to existing line segment detectors by simply replacing their feature extraction backbone. To verify the generality of the feature fusion network, we apply it to the current state-of-the-art line segment detectors HAWP \cite{xue2020holistically} and ULSD \cite{li2021ulsd}, named FE-HAWP and FE-ULSD respectively.

\section{Experiments and Analysis}
\label{sec:experiments}
In this section, we first detail the dataset, the experimental settings, and evaluation metrics. Then the significant component configurations of FE-LSD are analyzed. Finally, FE-LSD is evaluated and compared with the state-of-the-art. 

\subsection{Dataset}
LSD datasets such as Wireframe \cite{huang2018learning} and YorkUrban \cite{cho2017novel} play a vital role in LSD methods. However, there is no publicly available LSD dataset with geometrically and temporally aligned RGB images and events. Additionally, annotation from scratch on either motion-blurred RGB images or EST frames is difficult to obtain accurate line segment annotations. Based on these requirements, we first build a larger-scale synthetic dataset with the existing Wireframe dataset \cite{huang2018learning}. Then, considering there may be large gaps between synthetic and real data, we contribute a real dataset by collecting real data and manually labeling line segments. 
\begin{figure}[htb]
	\centering
	\vspace{-0.4cm}
	\includegraphics[width=0.99\linewidth]{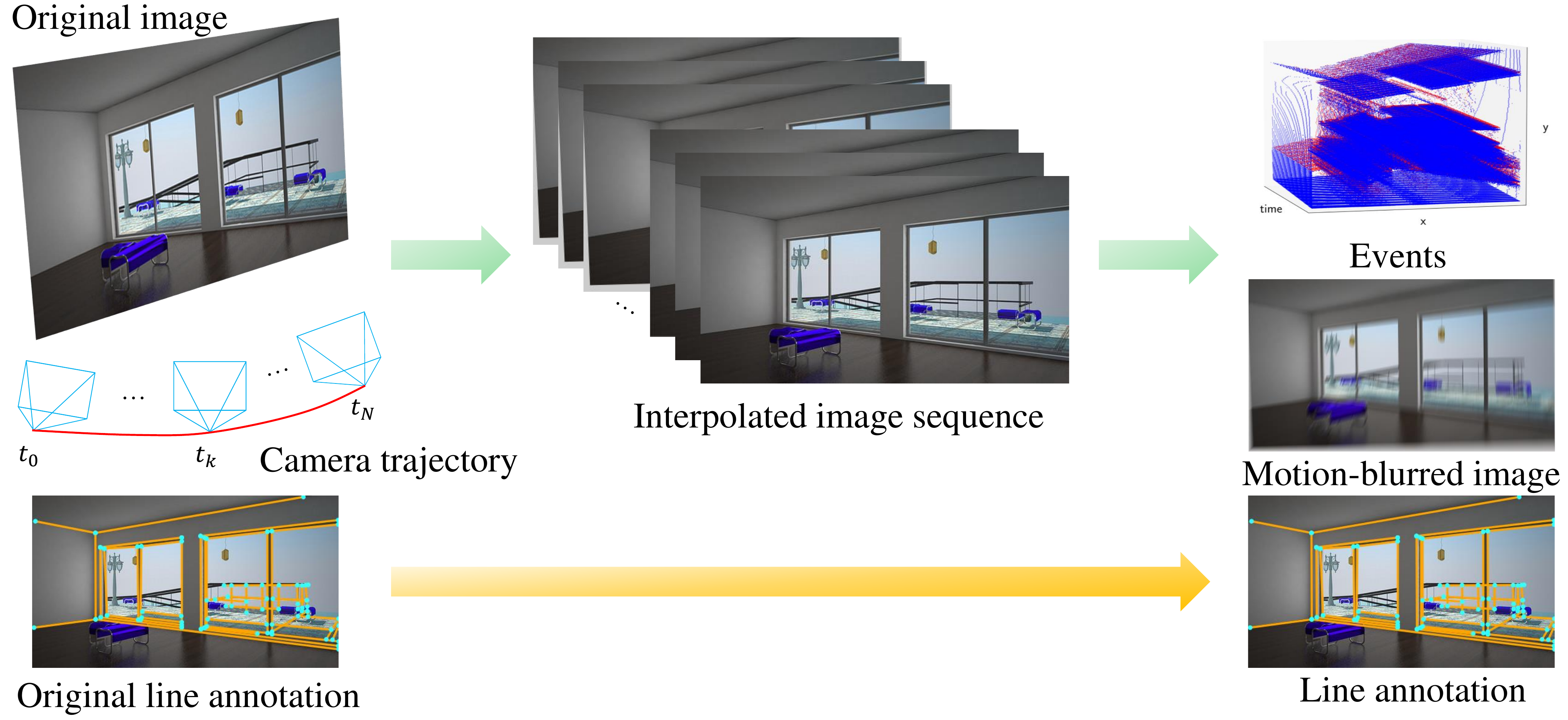}
	\caption{Synthetic LSD dataset generation with the registered motion-blurred RGB image and events.}
	\label{simu_data}
\end{figure}

\textbf{FE-Wireframe}:
The goal of synthetic data generation is to avoid laborious line annotation over motion blurs. With the labeled Wireframe dataset \cite{huang2018learning}, we use ESIM~\cite{rebecq_esim_2018} to generate the synthetic data, and the construction pipeline is shown in Fig.~\ref{simu_data}. Firstly, the RGB frame $I$ is re-projected to 3D space and then projected to image space to obtain frame $I_0$ using the initial camera pose and camera intrinsics. With the simulated camera trajectory over a short time window $T$, frame $I_k$ can be generated at any interpolation time $t_k$. Next, using the two adjacent frames $I_{k-1}$ and $I_{k}$ at time $t_{k-1}$ and $t_k$, we can obtain the intensity changes and use Eq. \ref{eq:eventgen} to generate the events $\mathcal{E}_k$. The time window $T$=$30$ ms, and the threshold $C$ for event contrast is randomly sampled from a uniform distribution $U(0.05, 0.5)$ for every image sequence. Then we can obtain the event data $\mathcal{E}=\bigcup_{k=0}^{N-1}\mathcal{E}_k$ and the simulated frames $\mathcal{I} = \left\{I_{k}\right\}_{k=0}^{N}$ for $N$+$1$ interpolated images. To directly use the original line annotation of the Wireframe dataset, the camera motion trajectory is designed to ensure that the last image $I_{N}$ is exactly the original image. The motion-blurred image $I_B$ is then obtained by averaging the $N$+$1$ images. Finally, we obtain the synthetic event data $\mathcal{E}$, the synchronized motion-blurred image $I_B$, and the line segment annotations. The FE-Wireframe dataset consists of 5000 pairs of frame-event training samples and 462 pairs of testing samples.

\textbf{\whu}:
For the generation of the real dataset with Frame-Event Line Segment detection in Motion-blur (\whu), we build a dual camera system similar to \cite{wang2020joint}, which consists of a DAVIS 346 event camera, FLIR RGB camera, and a beam splitter mounted in front of the two cameras. 52 data sequences are collected on the campus of Wuhan University, including typical indoor and outdoor scenes, such as libraries, dining rooms, gymnasiums, office buildings, and classrooms. Each data sequence includes an event stream and APS images from the event camera, and synchronized RGB images from the FLIR camera. The resolution of the event camera and FLIR RGB camera are $346$$\times$$260$ and $640$$\times$$512$, respectively. The frame rate for FLIR is set as 200 Hz to ensure that a clear image can be captured for line segment annotation. 

We crop 800 fragments from the 52 sequences, each fragment contains events over 30 ms and the synchronized 7 RGB frames. Then, the Super SloMo \cite{jiang2018super} algorithm is used to interpolate the RGB images from 7 to 40 frames for motion-blurred image generation.  For the annotation, we label on the last clear image of the 30ms interval. We refer to the clear image and event data to improve and validate the annotation quality. Finally, the real dataset contains 800 samples, including clear images, synthetic blurred images, event streams, and line segment annotation. The dataset is randomly divided into a training set with 600 samples and a test set with 200 samples.

\subsection{Experimental settings and evaluation metrics}
\textbf{Experimental setup.} In the experiment, the temporal dimension of EST $B$ is set to 5. The sizes of the image and EST are re-scaled to $512$$\times$$512$ before being input into the network. In the feature fusion backbone, the number of pairwise hourglass modules is set to 2, and the number of feature encoders in each module is set to 5. After the feature fusion backbone network, we obtain a fused feature map with the size of $128$$\times$$128$$\times$$256$. The parameters and the training configurations of FE-HAWP are also set to be the same as the original HAWP. The network is trained using the Adam optimizer \cite{kingma_adam_2015}. The learning rate, weight decay, and training batch size for network training are set to $4$$\times$$10^{-4}$, $1$$\times$$10^{-4}$, and 4, respectively. In the line proposal module, the numbers of junction proposals and line proposals are set as $K_{\mathrm{junc}}$=300, $K_{\mathrm{line}}$=5000. To train the line classifier, the numbers of positive and negative samples from LPN are set as $N_{\mathrm{pos}}$=300, $N_{\mathrm{neg}}$=300, respectively. The threshold of the binary classification confidence score is set the same as it is in HAWP \cite{xue2020holistically} and ULSD \cite{li2021ulsd}. Additionally, we add $N_{\mathrm{pos}}^*$=300 positive samples from ground truth line annotations to the training dataset. To obtain the feature vector for each line, 32 points are uniformly sampled from the line, and the dimension of the feature vector is set as 1024. Meanwhile, the training utilizes a step decay learning strategy, the network is trained with 30 epochs. After 25 epochs, the learning rate is divided by 10. The data argumentation used in this work includes horizontal and vertical flips, and $180\degree$ rotation. The model training and testing are conducted on a single NVIDIA RTX 3090 GPU.

\textbf{Evaluation metrics.} We use evaluation metrics commonly used in line segment detection tasks, including AP$^H$, F$^H$, mean structural Average Precision (msAP), sAP$^{5}$, sAP$^{10}$, sAP$^{15}$, junction mean Average Precision (mAP$^J$) and Frames Per Second (FPS). 

AP$^H$ and F$^H$ are calculated using precision and recall, which follow the traditional LSD \cite{von2010lsd} and L-CNN \cite{huang2018learning}. sAP is proposed in L-CNN \cite{zhou_end2019} to reflect the geometric structure. 
The distance between the predicted line and the GT line is used to determine whether a predicted line segment is a true positive. sAP$^{5}$, sAP$^{10}$, and sAP$^{15}$ are the sAP values when the distance is less than 5, 10, and 15 pixels, respectively. While msAP is the average of sAP$^{5}$, sAP$^{10}$, and sAP$^{15}$. For the distance calculation:
\begin{equation} 
	\label{eq:linedist}
	d(l,\hat{l}) = \frac{1}{2} \min \left(
	\sum_{i=0}^{1}\left\| \boldsymbol{p}_{i}-\boldsymbol{\hat{p}}_{i} \right\|,
	\sum_{i=0}^{1}\left\| \boldsymbol{p}_{i}-\boldsymbol{\hat{p}}_{1-i} \right\|
	\right),
\end{equation}
where $\{\boldsymbol{p}_{i}\}_{i=0}^1$ and $\{\boldsymbol{\hat{p}}_{i}\}_{i=0}^1$ are the start and end points of GT and predicted line segments. 

mAP$^J$ is used to measure the precision of junction predictions, which is computed in a similar way to the sAP of line segments. A true positive junction is determined by calculating the Euclidean distance between predicted and GT junctions. mAP$^J$ is the average accuracy when setting the distance thresholds as 0.5, 1, and 2 pixels. 


\subsection{Configuration analyses}
In this section, we first conduct experiments on the synthetic FE-Wireframe dataset to analyze the influence of the three main configurations, \ie, input data fusion, event representation, and fusion strategy. Then, the model trained on the FE-Wireframe dataset is fine-tuned on the real dataset to show the transfer learning results.

\begin{table}[htbp]
	\centering
	\caption{Impact of the input data fusion on network performance}
	\label{tab:input}
	\begin{threeparttable}
	\begin{tabular}{c|cc|ccccccc}
		\hline 
		 & \tiny{F} & \tiny{E} & \tiny{sAP$^5$} & \tiny{sAP$^{10}$} & \tiny{sAP$^{15}$} & \tiny{msAP} & \tiny{mAP$^J$} & \tiny{AP$^H$} & \tiny{F$^H$} \\ 
		\hline
        \multirow{3}*{\rotatebox{90}{\tiny{HAWP}}} & \checkmark &  & 41.6 & 47.3 & 49.9 & 46.3 & 44.0 & 73.5 & 72.2 \\
        &  & \checkmark & 25.3 & 29.7 & 32.0 & 29.0 & 31.2 & 54.4 & 59.1 \\
        & \checkmark & \checkmark & \textbf{45.1} & \textbf{50.4} & \textbf{52.9} & \textbf{49.5} & \textbf{46.8} & \textbf{75.0} & \textbf{73.2} \\
        \hline
        \multirow{3}*{\rotatebox{90}{\tiny{ULSD}}}  & \checkmark & & 40.8 & 46.7 & 49.4 & 45.6 & 45.1 & 67.5 & 70.6 \\
        &  & \checkmark & 25.9 & 30.6 & 33.1 & 29.8 & 32.6 & 48.4 & 59.2 \\
        & \checkmark & \checkmark & \textbf{47.0} & \textbf{52.7} & \textbf{55.2} & \textbf{51.7} & \textbf{48.8} & \textbf{72.2} & \textbf{73.7} \\
		\hline
	\end{tabular}
	\begin{tablenotes}
		\footnotesize 
		\item[1] F - image frames, E - events
	\end{tablenotes}
	\end{threeparttable}
\end{table}

\subsubsection{Input data fusion}
To analyze the effect of input data on network performance, we train on the blurred image only, the EST only, and the concatenation of both using the original HAWP \cite{xue2020holistically} and ULSD \cite{li2021ulsd} with the stacked hourglass network backbone. It should be noticed that the results on blurred images are retrained using blurred images. The results are shown in Table \ref{tab:input}. As for single-modal input, higher accuracy can be obtained using only image frames. This is because the event data loses a lot of texture information, and event noises have a negative impact on junction prediction. Most importantly, when using events to enhance image-based line segment detection, the fusion using naive concatenation can comprehensively improve the line detection performance. For example, the msAP of HAWP on frame-event concatenation is 3.2 and 20.5 points higher than the model trained with images and events, respectively. The ULSD model shows consistent improvements with 6.1 and 21.9 points higher msAP. These results demonstrate the benefits of edge-aware events over high-speed motion blurs. Thus simple concatenation can fuse the complementary information for line segment detection. 

\begin{table}[htpb]
	\centering
	\caption{Impact of event representation on network performance}
	\label{tab:eventImpac}
	\begin{tabular}{c|c|ccccccc}
		\hline
		 & \tiny{Event} & \tiny{sAP$^5$} & \tiny{sAP$^{10}$} & \tiny{sAP$^{15}$} & \tiny{msAP} & \tiny{mAP$^J$} & \tiny{AP$^H$} & \tiny{F$^H$} \\
		\hline
		\multirow{4}*{\rotatebox{90}{\tiny{FE-HAWP}}} & \tiny{EST} & \textbf{48.7} & \textbf{53.9} & \textbf{56.2} & \textbf{53.0} & \textbf{49.4} & \textbf{77.1} & 75.1 \\
        & \tiny{Voxel Grid} & 48.3 & 53.5 & 55.8 & 52.6 & 48.9 & 76.9 & \textbf{75.3} \\
        & \tiny{EC+SAE} & 48.3 & 53.7 & 55.8 & 52.6 & 49.1 & 77.0 & 75.1 \\
        & \tiny{EC+SAE*} & 47.3 & 52.5 & 54.8 & 51.5 & 47.8 & 76.0 & 74.1 \\
		\hline	
		\multirow{4}*{\rotatebox{90}{\tiny{FE-ULSD}}} & \tiny{EST} & \textbf{50.9} & \textbf{56.5} & \textbf{58.8} & \textbf{55.4} & \textbf{51.1} & \textbf{75.3} & \textbf{75.9} \\
        & \tiny{Voxel Grid} & 50.3 & 55.7 & 58.0 & 54.7 & 50.9 & 74.8 & 75.3 \\
        & \tiny{EC+SAE} & 50.7 & 56.1 & 58.3 & 55.0 & 50.9 & 74.3 & \textbf{75.5} \\
        & \tiny{EC+SAE*} & 49.4 & 55.0 & 57.3 & 53.9 & 50.4 & 74.0 & 74.7 \\
		\hline
	\end{tabular}
\end{table}

\subsubsection{Event representation}
Different event representation determines the richness of the event information, directly affecting the performance of FE-LSD. To investigate the impact, four different event representations, \ie, Event Spike Tensor (EST) \cite{gehrig_end--end_2019}, voxel grid \cite{zhu_unsupervised_2019}, Event Count + Surface of Active Event(EC+SAE) \cite{maqueda2018event, benosman2013event}, Event Count + Surface of Active Event without polarity (EC+SAE$^*$), are tested. The channels of event frames are 10, 5, 4, and 2 for the four representations, respectively. Experimental results are shown in Table \ref{tab:eventImpac}. Among the four representations, when adding polarity information to the Voxel Grid and EC+SAE$^*$, \ie, EST and EC+SAE, the comprehensive performance is improved for all metrics. 
Besides, EST and Voxel Grid retain the temporal information compared with EC+SAE and EC+SAE$^*$, therefore obtaining much better accuracy than the other two representations, indicating the importance of the temporal information for the motion-blurred line segment detection task. As for the pixel-based metrics AP$^H$ and F$^H$, neither of them is affected by the event representations, indicating that polarity and temporal information mainly enhance the accuracy of structured line detection. Most importantly, the EST-based line detectors have the highest accuracy, which encodes the polarity information and retains the most temporal information. Therefore, EST is selected as the final event representation of FE-LSD.

\begin{table}[htbp]
	\centering
	\caption{Impact of the fusion module on network performance}
	\label{tab:fusion}
	\begin{tabular}{c|cc|ccccccc}
		\hline
		 & \tiny{SFB} & \tiny{DFB} & \tiny{sAP$^5$} & \tiny{sAP$^{10}$} & \tiny{sAP$^{15}$} & \tiny{msAP} & \tiny{mAP$^J$} & \tiny{AP$^H$} & \tiny{F$^H$} \\
		\hline
		\multirow{4}*{\rotatebox{90}{\tiny{FE-HAWP}}} &  &  & 43.6 & 49.1 & 51.5 & 48.1 & 45.8 & 74.5 & 73.5 \\
        & \checkmark &  & 46.0 & 51.2 & 53.4 & 50.2 & 47.6 & 75.5 & 73.3 \\
        &  & \checkmark & 48.5 & 53.7 & 55.9 & 52.7 & 48.9 & 76.8 & 74.6 \\
        & \checkmark & \checkmark & \textbf{48.7} & \textbf{53.9} & \textbf{56.2} & \textbf{53.0} & \textbf{49.4} & \textbf{77.1} & \textbf{75.1} \\
		\hline	
		\multirow{4}*{\rotatebox{90}{\tiny{FE-ULSD}}} &  &  & 45.5 & 51.2 & 53.7 & 50.1 & 47.6 & 71.7 & 72.8 \\
        & \checkmark &  & 46.5 & 52.2 & 54.5 & 51.1 & 48.9 & 71.1 & 73.2 \\
        &  & \checkmark & 50.1 & 55.6 & 57.9 & 54.5 & 50.5 & \textbf{74.5} & 75.0 \\
        & \checkmark & \checkmark & \textbf{50.9} & \textbf{56.5} & \textbf{58.8} & \textbf{55.4} & \textbf{51.1} & \textbf{75.3} & \textbf{75.9} \\
		\hline
	\end{tabular}
\end{table}
 \subsubsection{Fusion module analysis}
For the fusion of image and event features, we test the effect of the proposed SFB and DFB, and the results are shown in Table \ref{tab:fusion}. The SFB is directly removed from the network when unused, and the element-by-element addition is used when the DFB is unused. From Table \ref{tab:fusion}, we can see that the network has the lowest accuracy when no fusion module is used, with 48.1$\%$ and 50.1$\%$ msAP for FE-HAWP and FE-ULSD, respectively. Then, msAP is improved by 2.1 and 1.0 points when the SFB is added, and it increases by 4.6 and 4.4 points when only the DFB is used, for FE-HAWP and FE-ULSD, respectively. The gains brought by DFB are larger than SFB, indicating the importance of the multi-scale encoder-decoder fusion of deep features. Therefore, the global attention in Transformer is helpful for line segment detection. When both SFB and DFB are used, we obtain the highest accuracy with 53.2$\%$ and 55.2$\%$ msAP for FE-HAWP and FE-ULSD, respectively. The above quantitative results verify the effectiveness of the proposed SFB and DFB.

\begin{figure}[htbp]
	\centering
	\includegraphics[width=\linewidth]{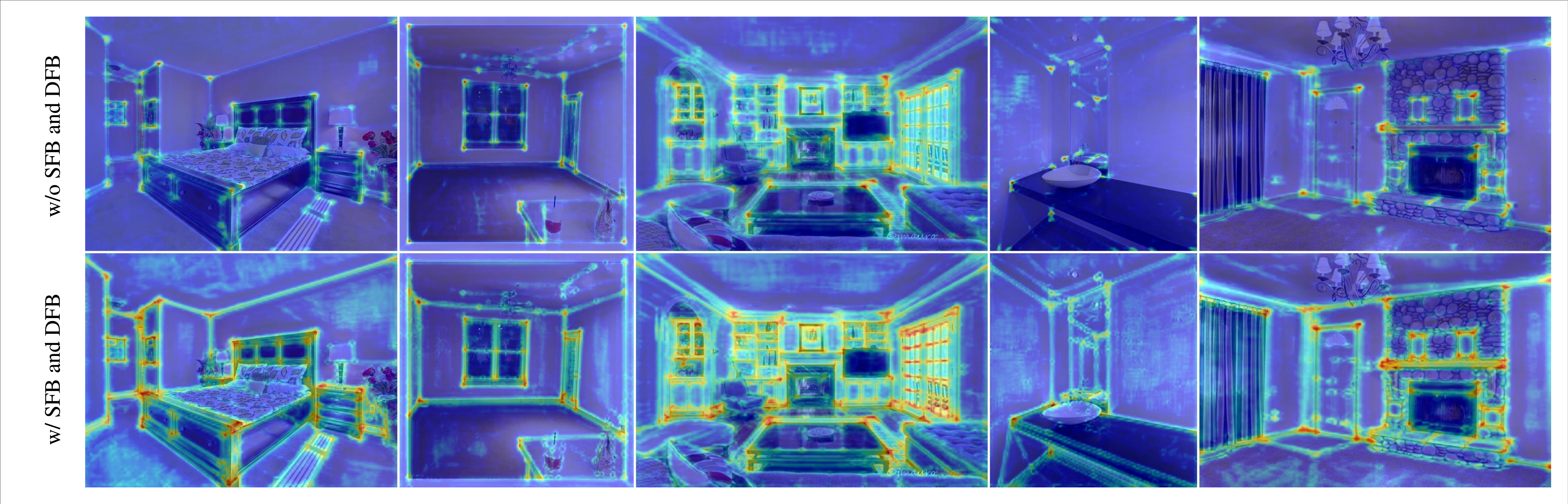}
	\caption{Comparison of Class Activation Map (CAM) visualization results on the synthetic dataset.}
	\label{fig:cam}
\end{figure}
Meanwhile, we further qualitatively analyze the effectiveness of the two fusion blocks by visualizing the feature map using Class Activation Map (CAM) \cite{even2019thick}. The results are shown in Fig. \ref{fig:cam}. The first row shows some sample CAMs trained using FE-HAWP without SFB and DFB, while the second row shows the corresponding samples using FE-HAWP with both modules. The CAM reflects the network attention distribution to different regions in the image, and the red indicates greater attention to these parts. By adding the SFB and DFB, it is clear that the network pays more attention to the edges and has stronger responses to the endpoints of line segments. These visualizations validate the importance of the SFB and DFB modules in improving the network's ability to recognize and locate line segments.

\subsubsection{Transfer learning on real data} 
In our experiment, the synthetic data generated on the Wireframe dataset relieves the burden of extensive data annotation on real images and events. However, the model trained on synthetic data has generalization issues to the real-world dataset. Therefore, we build a real dataset \whu to conduct transfer learning. To verify the effectiveness of transfer learning, the three models, \ie, trained on synthetic only, trained on real only, and pre-trained on synthetic and fine-tuned on real, are evaluated on the same real dataset. The results are shown in Table \ref{tab:model-transfer}. Compared with real data training (R) and fused data training (F), it can be observed that the models trained on synthetic only (S) obtain $30.7\%$ and $34.4\%$ msAP on the real-world testing set for FE-HAWP and FE-ULSD, respectively. This performance degradation is similar to the HAWP \cite{xue2020holistically} when it is trained on Wireframe only and tested on the YorkUrban dataset. The reasons are mainly two-fold: (\romannumeral1) The significant data differences between synthetic and real datasets; (\romannumeral2) The synthetic event data is obtained based on the ideal event generation model, while the quality of real event data is affected by the lens, noise, contrast threshold and other factors for event cameras. Then comparing the transfer learning models (F) with the models trained on synthetic (S) and real only (R), the performance of the synthetically pre-trained model has improved substantially after being fine-tuned on the real dataset and also has a significant improvement over the model trained on real only. The msAPs are finally improved to $63.3\%$ and $62.9\%$ for the FE-HAWP and FE-ULSD models, respectively.
\begin{table}[htbp]
	\centering
	\caption{Effect of model transfer and fine-tuning on real dataset}
	\label{tab:model-transfer}
	\begin{threeparttable}
		\begin{tabular}{c|c|ccccccc}
			\hline
			 & \tiny{Train} & \tiny{sAP$^5$} & \tiny{sAP$^{10}$} & \tiny{sAP$^{15}$} & \tiny{msAP} & \tiny{mAP$^J$} & \tiny{AP$^H$} & \tiny{F$^H$} \\
			\hline
			\multirow{3}*{\rotatebox{90}{\tiny{FE-HAWP}}} & S & 26.8 & 31.4 & 33.7 & 30.7 & 30.1 & 45.9 & 54.9 \\
            & R & 47.5 & 53.0 & 55.4 & 52.0 & 50.9 & 74.0 & 73.9 \\
            & F & \textbf{59.8} & \textbf{64.2} & \textbf{65.9} & \textbf{63.3} & \textbf{60.1} & \textbf{82.0} & \textbf{79.7} \\
			\hline
			\multirow{3}*{\rotatebox{90}{\tiny{FE-ULSD}}} & S & 30.4 & 35.3 & 37.6 & 34.4 & 31.4 & 48.0 & 62.5 \\
            & R & 47.3 & 52.9 & 55.2 & 51.8 & 52.2 & 72.9 & 73.7 \\
            & F & \textbf{59.3} & \textbf{63.8} & \textbf{65.7} & \textbf{62.9} & \textbf{61.0} & \textbf{77.8} & \textbf{77.1} \\
			\hline	
		\end{tabular}
		\begin{tablenotes}
			\footnotesize 
			\item[-] S-training on the synthetic only, R - training on the real only, F - pre-training on the synthetic  and fine-tuning on real.
		\end{tablenotes}
	\end{threeparttable}
\end{table}

\begin{figure}[htb]
	\begin{minipage}[t]{0.495\linewidth}
		\centering
		\includegraphics[width=\columnwidth]{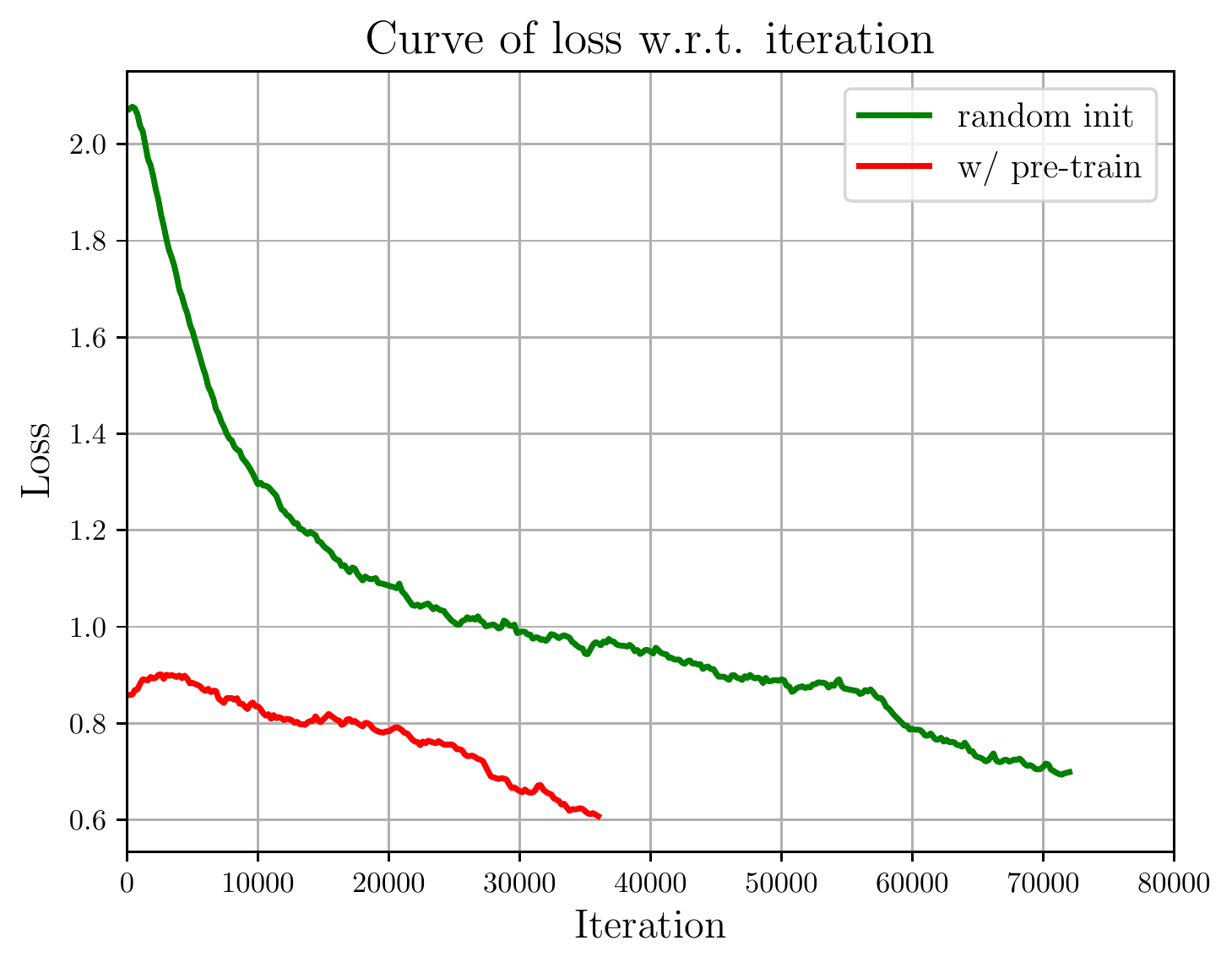} 
	\end{minipage}
	\begin{minipage}[t]{0.495\linewidth} 
		\centering
		\includegraphics[width=\columnwidth]{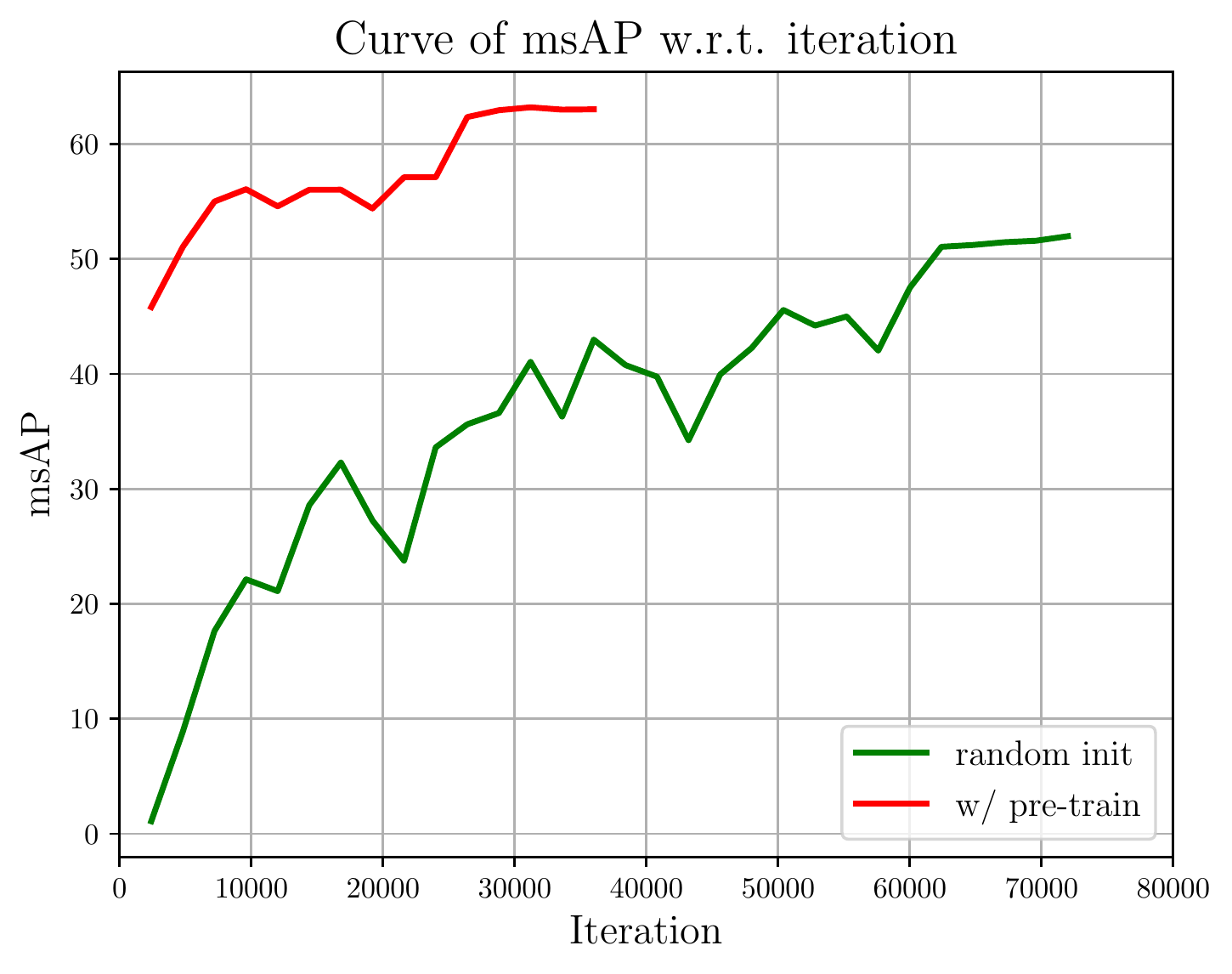} 
	\end{minipage}
	\caption{Fine-tuning impact on the training process}
	\label{fig:finetuning}
\end{figure}

We further plot the curves of loss and $msAP$ in Fig. \ref{fig:finetuning}. The red curves are trained from the pre-trained model, and the green curves are trained from scratch using random initialization. It is clear that fine-tuning the pre-trained model can speed up the convergence and improve the final performance. These results indicate a notable difference between the synthetic and the real-world dataset, which seriously affects the model's generalization ability when evaluating from simulation to the real world. Nevertheless, the large-scale synthetic dataset is helpful for model pre-training, thus releasing the data annotation labor and improving the convergence speed and accuracy on the real dataset.

\subsection{Main results for comparison}
To evaluate the performance of the proposed FE-LSD, sufficient comparisons are conducted on the synthetic FE-Wireframe and real \whu ~datasets, respectively. The competitors include traditional LSD~\cite{von2010lsd}, FBSD~\cite{even2019thick} (designed for blurred images), the learning-based methods L-CNN~\cite{zhou_end2019}, HAWP~\cite{xue2020holistically}, LETR \cite{xu_line_2021}. These learning-based methods are directly evaluated using the officially trained models on the motion-blurred images. For FE-Wireframe, we additionally use the official HAWP and ULSD models on the deblurred images with RED-Net \cite{xu2021motion}, \ie, HAWP$^\dag$ and ULSD$^\dag$. Finally, since the proposed FE-LSD takes both image and event data inputs, these methods are retrained on concatenating images and EST to give fair play (\ie, L-CNN$^\ddag$, HAWP$^\ddag$, ULSD$^\ddag$ and LETR$^\ddag$).

\begin{figure*}[htbp]
	\centering
	\includegraphics[width=0.90\linewidth]{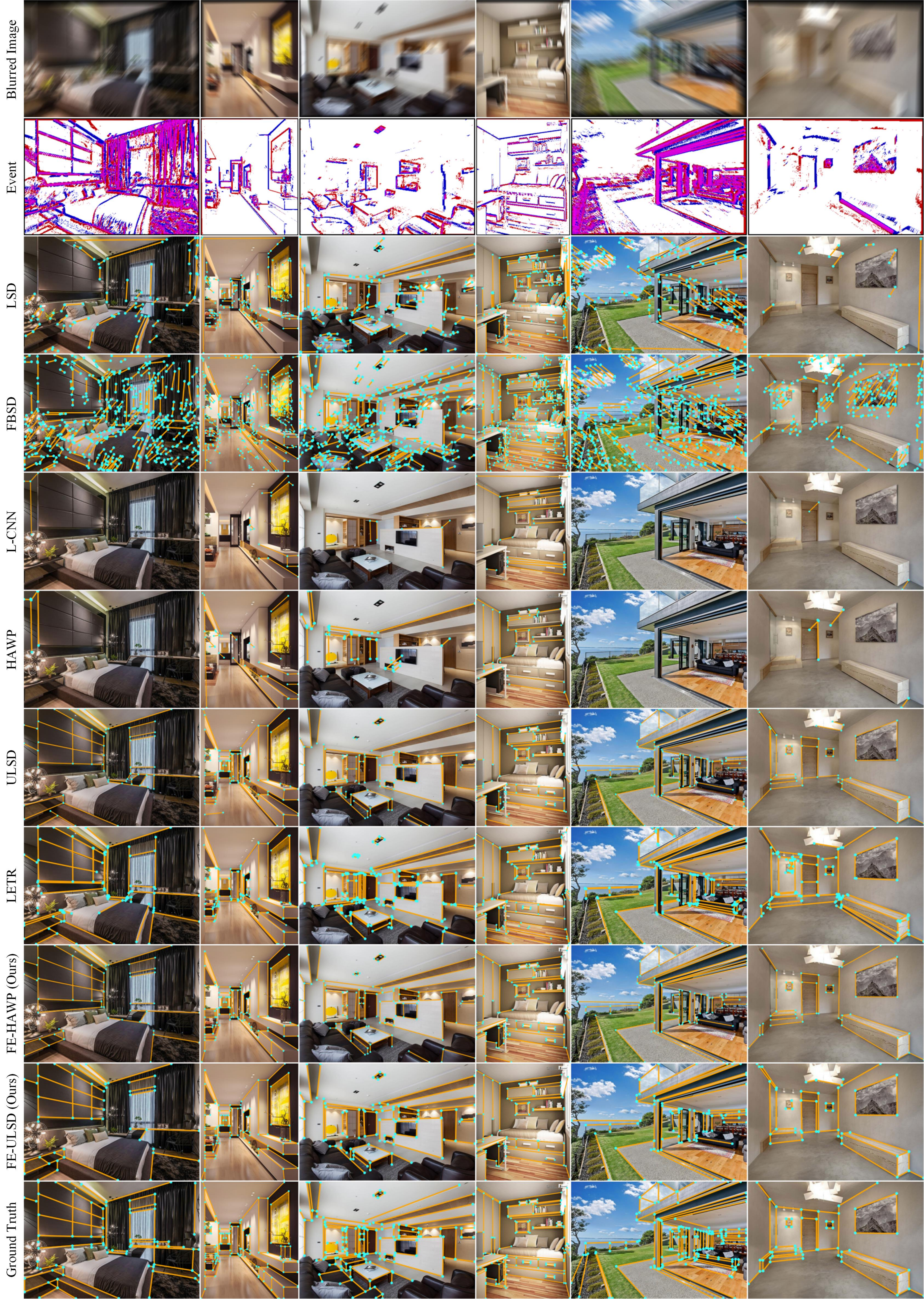}
	\caption{Result comparisons on FE-Wireframe dataset (Lines are plotted on clear images but detected on the motion-blurred images and events)}
	\label{fig:syn_compare}
\end{figure*}

\subsubsection{Results on the FE-Wireframe dataset}
Fig. \ref{fig:syn_compare} shows the qualitative comparisons on the FE-Wireframe dataset. The first row is the input motion-blurred images, and the second is the corresponding event inputs. For better result visualization, the detected lines and the manually-labeled lines are plotted on the clear RGB image at the end moment of exposure. Since LSD and FBSD are gradient-based methods, they can detect a large number of segments in motion-blurred images. However, the majority of the line segments are false detection or fragmented segments, resulting in low quantitative metrics. On the other hand, the learning-based methods HAWP and LETR have a low detection rate on blurry images when using the pre-trained models. This is because the officially trained models are driven by clear images and cannot handle the motion-blur issues. However, when we retrain them using the concatenation of blurry images and events, the correct detection is significantly increased. Finally, the proposed FE-HAWP and FE-ULSD have the best performance with the highest detection rate and the lowest false alarm compared to the ground truth.

\begin{table}[ht]
	\centering
	\caption{Quantitative comparisons on synthetic FE-Wireframe dataset}
	\label{tab:synth_result}
	\begin{tabular}{l|cccccc|c}
		\hline
		\tiny{Method} & \tiny{sAP$^5$} & \tiny{sAP$^{10}$} & \tiny{sAP$^{15}$} & \tiny{msAP} & \tiny{mAP$^J$} & \tiny{AP$^H$} & \tiny{FPS} \\ 
		\hline
		LSD\cite{von2010lsd} & 0.1 & 0.6 & 1.1 & 0.6 & 3.0 & 19.5 & \textbf{76.7} \\
		FBSD\cite{even2019thick} & 0.2 & 0.4 & 0.9 & 0.5 & 2.9 & 24.9 & 21.7 \\
		\hline
		L-CNN\cite{zhou_end2019} & 3.4 & 5.1 & 6.2 & 4.9 & 7.0 & 22.7 & 28.8 \\
		HAWP\cite{xue2020holistically} & 3.5 & 5.1 & 6.3 & 5.0 & 6.8 & 21.7 & 36.6 \\
		HAWP$^\dag$ & 7.0 & 9.8 & 11.2 & 9.3 & 9.3 & 24.8 & 39.7 \\
		ULSD \cite{li2021ulsd} & 3.5 & 5.3 & 6.8 & 5.2 & 7.5 & 20.2 & 39.7 \\
		ULSD$^\dag$ & 6.8 & 9.8 & 11.6 & 9.4 & 10.6 & 16.6 & 41.0 \\
		LETR\cite{xu_line_2021} & 2.8 & 5.0 & 6.5 & 4.8 & 7.3 & 21.9 & 4.2 \\
		\hline
		L-CNN$^{\ddag}$ & 40.6 & 45.8 & 48.2 & 44.8 & 45.6 & 70.5 & 10.6 \\
        HAWP$^{\ddag}$ & 45.1 & 50.4 & 52.9 & 49.5 & 46.8 & 75.0 & 26.8 \\
        ULSD$^{\ddag}$ & 47.0 & 52.7 & 55.2 & 51.7 & 48.8 & 72.2 & 32.2 \\
        LETR$^{\ddag}$ & 24.7 & 34.7 & 39.7 & 33.1 & 25.4 & 66.1 & 3.9 \\
		\hline
        FE-HAWP & 48.7 & 53.9 & 56.2 & 53.0 & 49.4 & \textbf{77.1} & 13.1 \\
        FE-ULSD & \textbf{50.9} & \textbf{56.5} & \textbf{58.8} & \textbf{55.4} & \textbf{51.1} & 75.3 & 13.9 \\
		\hline	
	\end{tabular}
	\begin{tablenotes}
		\footnotesize 
		\item[1] ${\dag}$ deblur first using \cite{xu2021motion} and detect line segments using the corresponding official models.
		\item[1] ${\ddag}$ retrain with the concatenation of image and EST.
	\end{tablenotes}
\end{table}

\begin{figure}[htbp]
    \centering
	\begin{minipage}[t]{0.495\linewidth}
		\centering
		\includegraphics[width=\columnwidth]{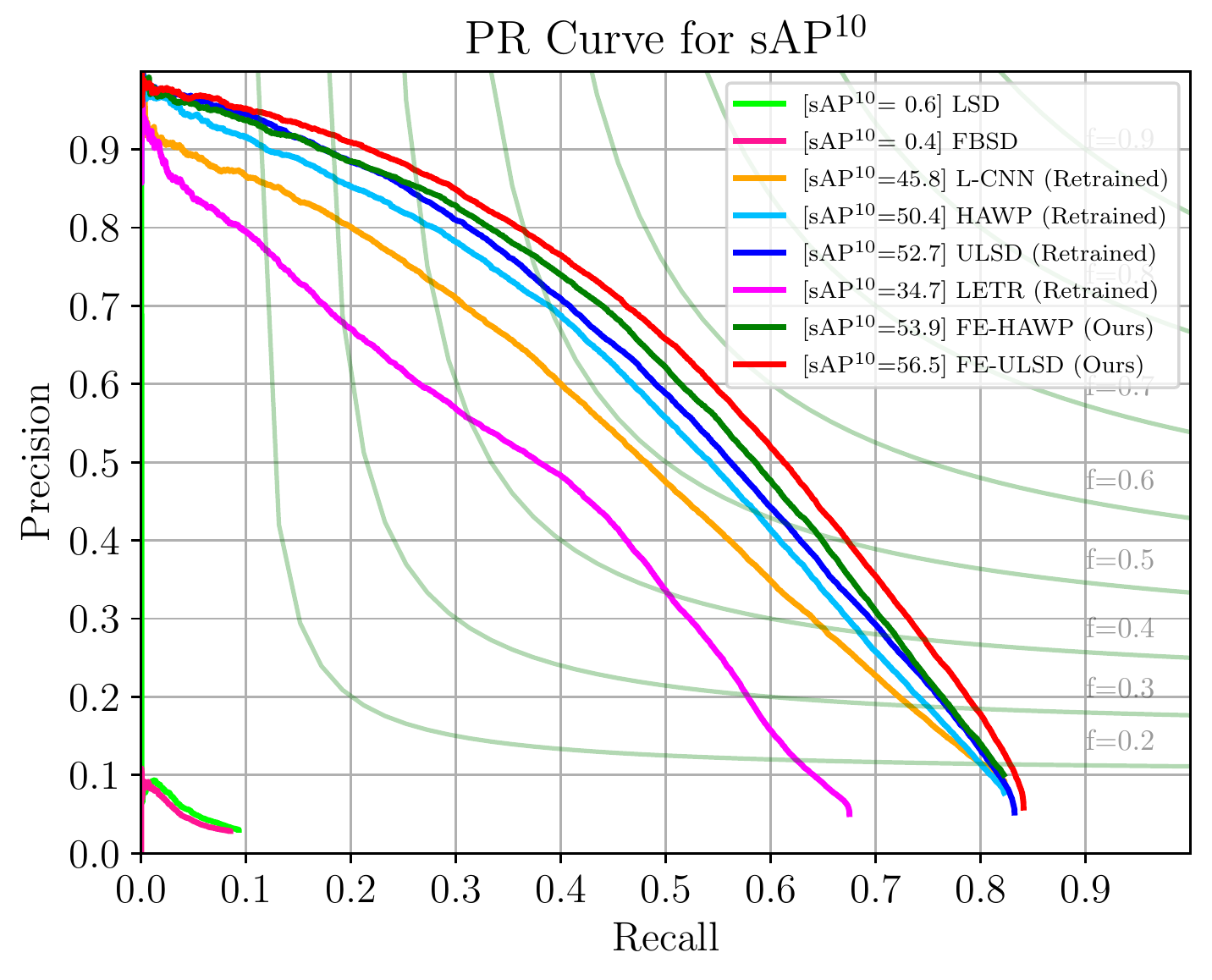}
	\end{minipage}
	\begin{minipage}[t]{0.495\linewidth} 
		\centering
		\includegraphics[width=\columnwidth]{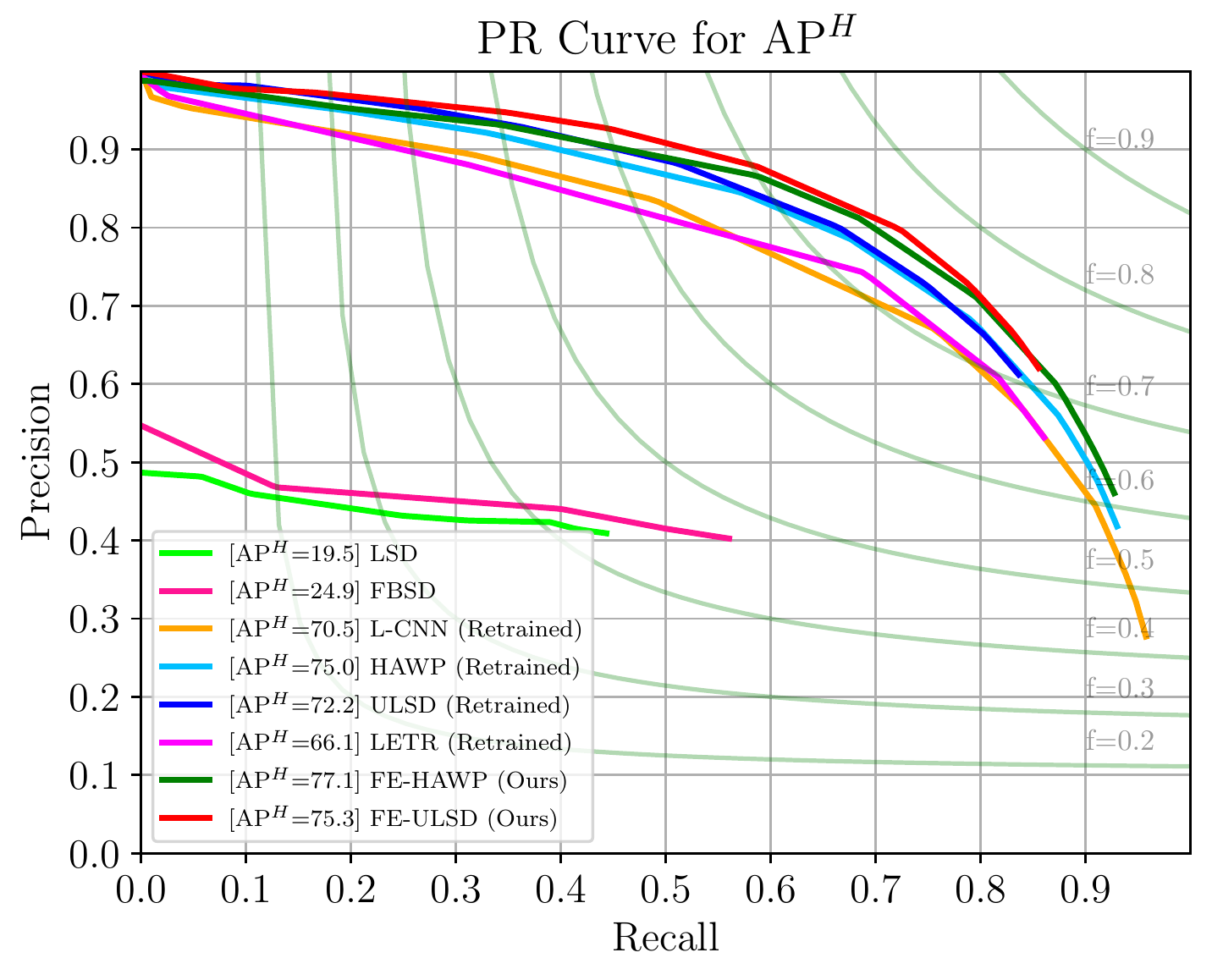} 
	\end{minipage}
	\caption{PR curves of sAP$^{10}$ and AP$^H$ on the FE-Wireframe dataset}
	\label{fig:prcurve_syn}
\end{figure}

Table \ref{tab:synth_result} summarizes the quantitative comparisons on the synthetic FE-Wireframe dataset, while Fig.~\ref{fig:prcurve_syn} shows the PR curve of sAP$^{10}$ and AP$^H$, respectively. When only given the motion-blurred images, the traditional LSD, FBSD, and deep learning-based L-CNN, HAWP, ULSD and HAWP have low accuracy. The highest msAP is obtained by ULSD for $5.2\%$, and FBSD obtains the highest AP$^H$ for $24.9\%$, respectively. Image deblur with events can improve the subsequent line segment detection performance. However, these results are far from the standard for SLAM and 3D reconstruction. Then, when taking both image and EST inputs, the retrained models perform much better than only blurry image input. The highest msAP is obtained by the retrained ULSD$\ddag$ for $51.7\%$. These improvements demonstrate that the event data can enhance the edge awareness for motion-blurred images. Finally, we obtain the highest performance for all metrics using our proposed frame-event fusion module. The msAPs of FE-HAWP and FE-USLD are higher than the retrained ULSD$\ddag$ for 1.3 and 3.7 points, respectively. Except for AP$^H$, FE-ULSD obtains the highest accuracy, while AP$^H$ is slightly lower than FE-HAWP. This is caused by the low recall rate of FE-ULSD when the confidence threshold is set to 0.5, as shown in the AP$^H$ PR curves. 

Regarding to the detection speed, both FE-ULSD and FE-HAWP are slower than the original methods because they use a feature fusion backbone with one more encoder branch than the original stacked hourglass network and introduce the more time-consuming Transformer for feature fusion. Nevertheless, compared with the transformer-based method LETR \cite{xu_line_2021}, the proposed method is nearly three times faster. These results prove the effectiveness of the proposed method.

\subsubsection{Results on \whu ~dataset}

\begin{figure}[htbp]
    \centering
	\begin{minipage}[t]{0.495\linewidth}
		\centering
		\includegraphics[width=\columnwidth]{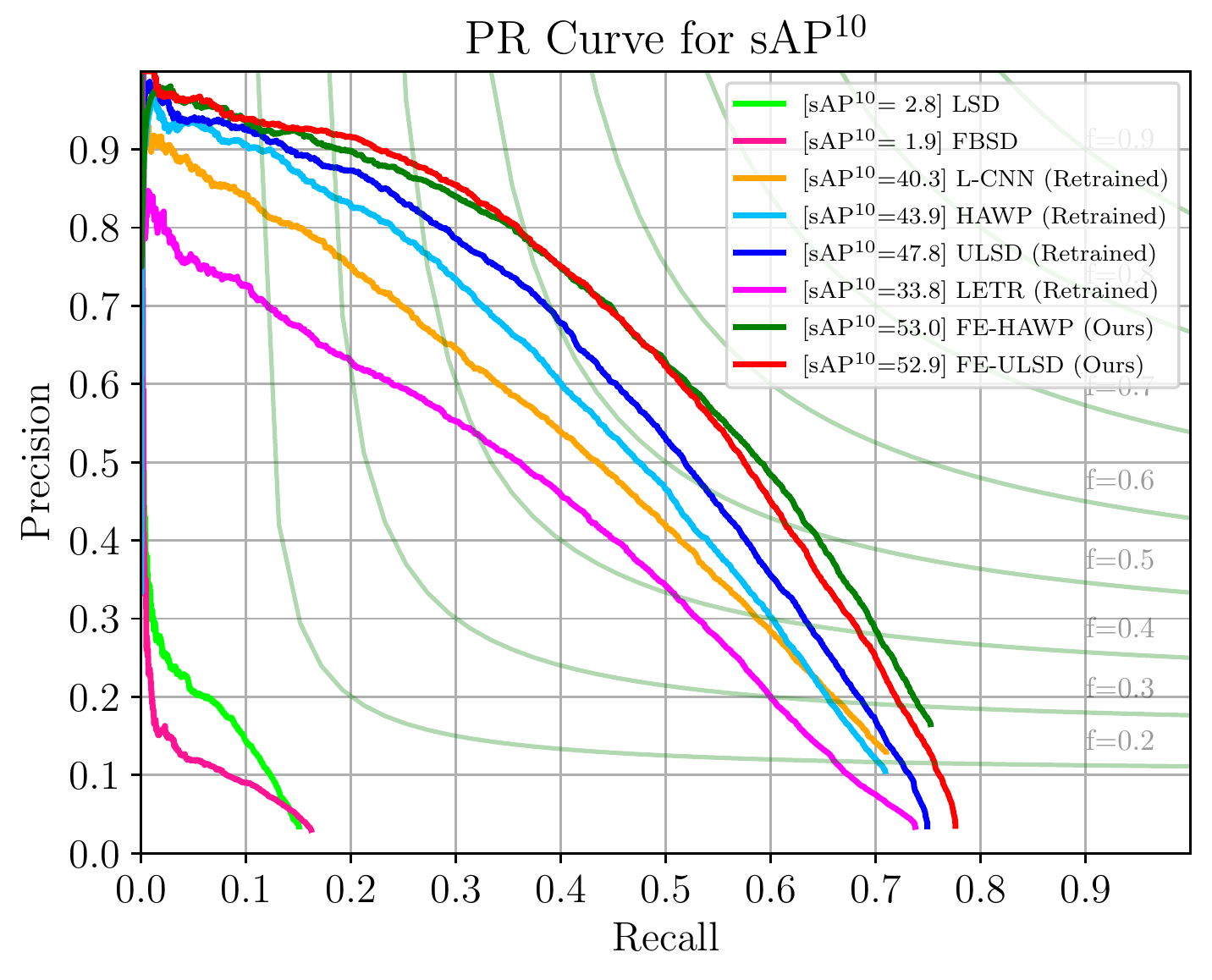}
	\end{minipage}
	\begin{minipage}[t]{0.495\linewidth} 
		\centering
		\includegraphics[width=\columnwidth]{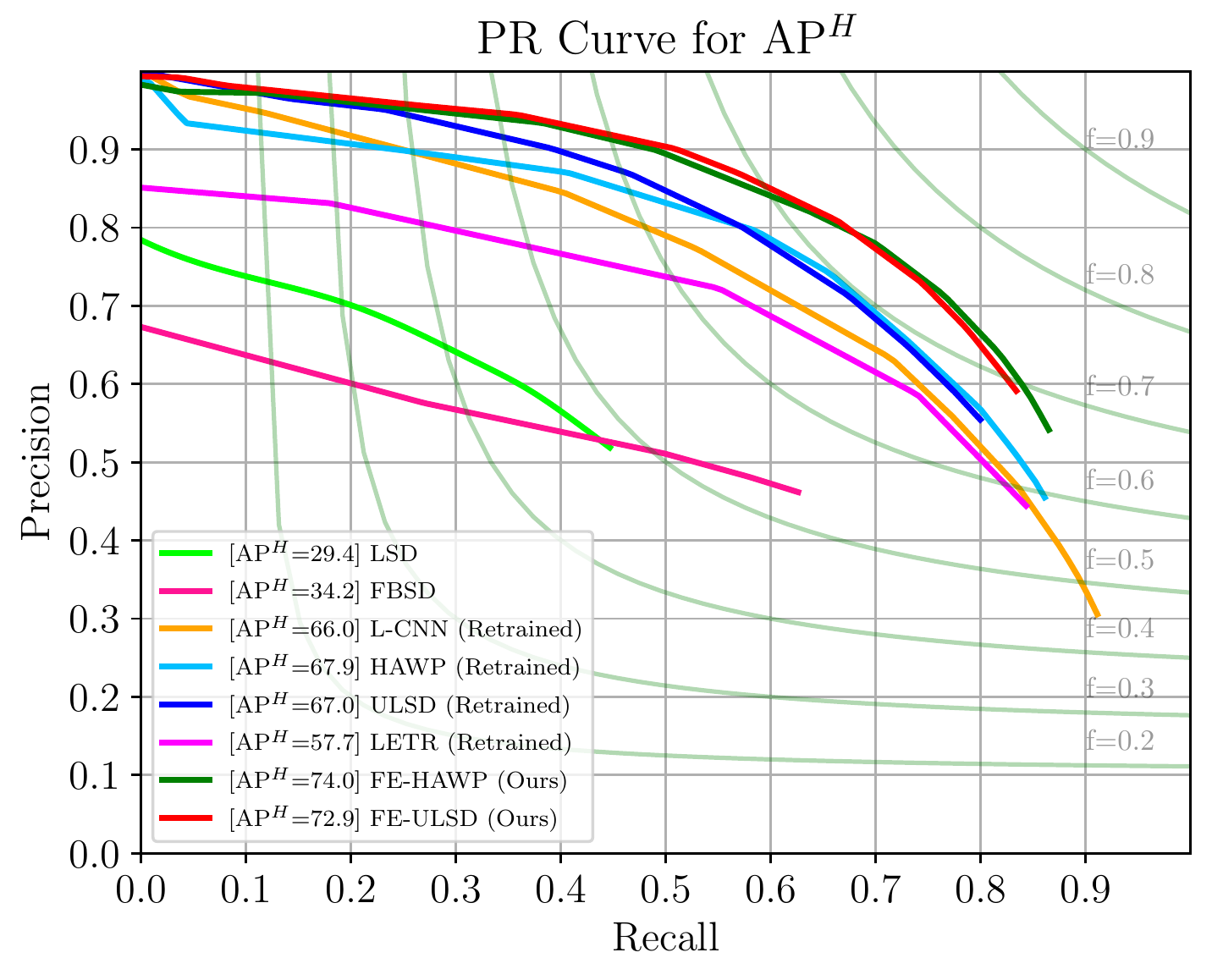}
	\end{minipage}
	\caption{PR curves of sAP$^{10}$ and AP$^H$ on \whu.}
	\label{fig:pr_real}
\end{figure}

Table \ref{tab:quati_real} and Fig. \ref{fig:pr_real} give the quantitative comparisons on the collected \whu ~dataset. The proposed FE-HAWP and FE-ULSD still achieve the best accuracy compared with other learning-based methods. FE-HAWP obtains $52.0\%$ msAP, $9.4$ points higher than the retrained HAWP$\dag$. FE-ULSD obtains $51.8\%$ msAP, $5.1$ points higher than the retrained ULSD$\dag$. Because the Bezier equipartition representation of line segments in ULSD is more sensitive to image noises than the attraction field map representation in HAWP, the gain of FE-ULSD is not as significant as FE-HAWP for real images. Furthermore, if we use the transfer learning strategy of fine-tuning on the pre-trained models, \ie, FE-HAWP$^{\ddag}$ and FE-ULSD$^{\ddag}$, the performances are improved by $11.3$ and $11.1$ points compared with FE-HAWP and FE-ULSD, respectively. 

\begin{table}[htbp]
	\centering
	\caption{Quantitative comparisons on the \whu~ dataset }
	\label{tab:quati_real}
	\begin{tabular}{l|cccccc|c}
		\hline
		\tiny{Method} & \tiny{sAP$^5$} & \tiny{sAP$^{10}$} & \tiny{sAP$^{15}$} & \tiny{msAP} & \tiny{mAP$^J$} & \tiny{AP$^H$}  & \tiny{FPS} \\
		\hline
		LSD\cite{von2010lsd} & 1.1 & 2.8 & 4.1 & 2.7 & 5.1 & 29.4 & \textbf{61.0} \\
		FBSD\cite{even2019thick} & 0.9 & 1.9 & 2.7 & 1.8 & 5.1 & 34.2 & 15.9 \\
		\hline
		L-CNN\cite{zhou_end2019} & 7.5 & 11.5 & 13.7 & 10.9 & 12.4 & 27.9 & 29.7 \\
		HAWP\cite{xue2020holistically} & 8.4 & 12.8 & 15.3 & 12.2 & 12.4 & 32.0 & 38.1 \\
		ULSD \cite{li2021ulsd} & 6.8 & 10.8 & 13.0 & 10.2 & 11.8 & 26.7 & 40.6 \\
		LETR\cite{xu_line_2021} & 7.1 & 13.0 & 16.8 & 12.3 & 12.1 & 30.2 & 3.6 \\
		\hline
		L-CNN$^{\dag}$ & 34.0 & 40.3 & 43.0 & 39.1 & 40.3 & 66.0 & 17.7 \\
        HAWP$^{\dag}$ & 37.0 & 43.9 & 46.9 & 42.6 & 41.6 & 67.9 & 29.0 \\
        ULSD$^{\dag}$ & 42.0 & 47.8 & 50.4 & 46.7 & 48.5 & 67.0 & 32.2 \\
        LETR$^{\dag}$ & 22.6 & 33.8 & 38.8 & 31.7 & 23.2 & 57.7 & 3.3 \\
		\hline	
        FE-HAWP & 47.5 & 53.0 & 55.4 & 52.0 & 50.9 & 74.0 & 12.9 \\
        FE-ULSD & 47.3 & 52.9 & 55.2 & 51.8 & 52.2 & 72.9 & 12.9 \\
        FE-HAWP$^{\ddag}$ & \textbf{59.8} & \textbf{64.2} & \textbf{65.9} & \textbf{63.3} & 60.1 & \textbf{82.0} & 13.3 \\
        FE-ULSD$^{\ddag}$ & 59.3 & 63.8 & 65.7 & 62.9 & \textbf{61.0} & 77.8 & 13.9 \\
		\hline	
	\end{tabular}
	\begin{tablenotes}
		\footnotesize 
		\item[1] ${\dag}$ retrain with the concatenation of image and EST.
		\item[1] ${\ddag}$ pre-train on FE-Wireframe and fine-tune on \whu.
	\end{tablenotes}
\end{table}

\begin{figure*}[!t]
	\centering
	\includegraphics[width=0.90\linewidth]{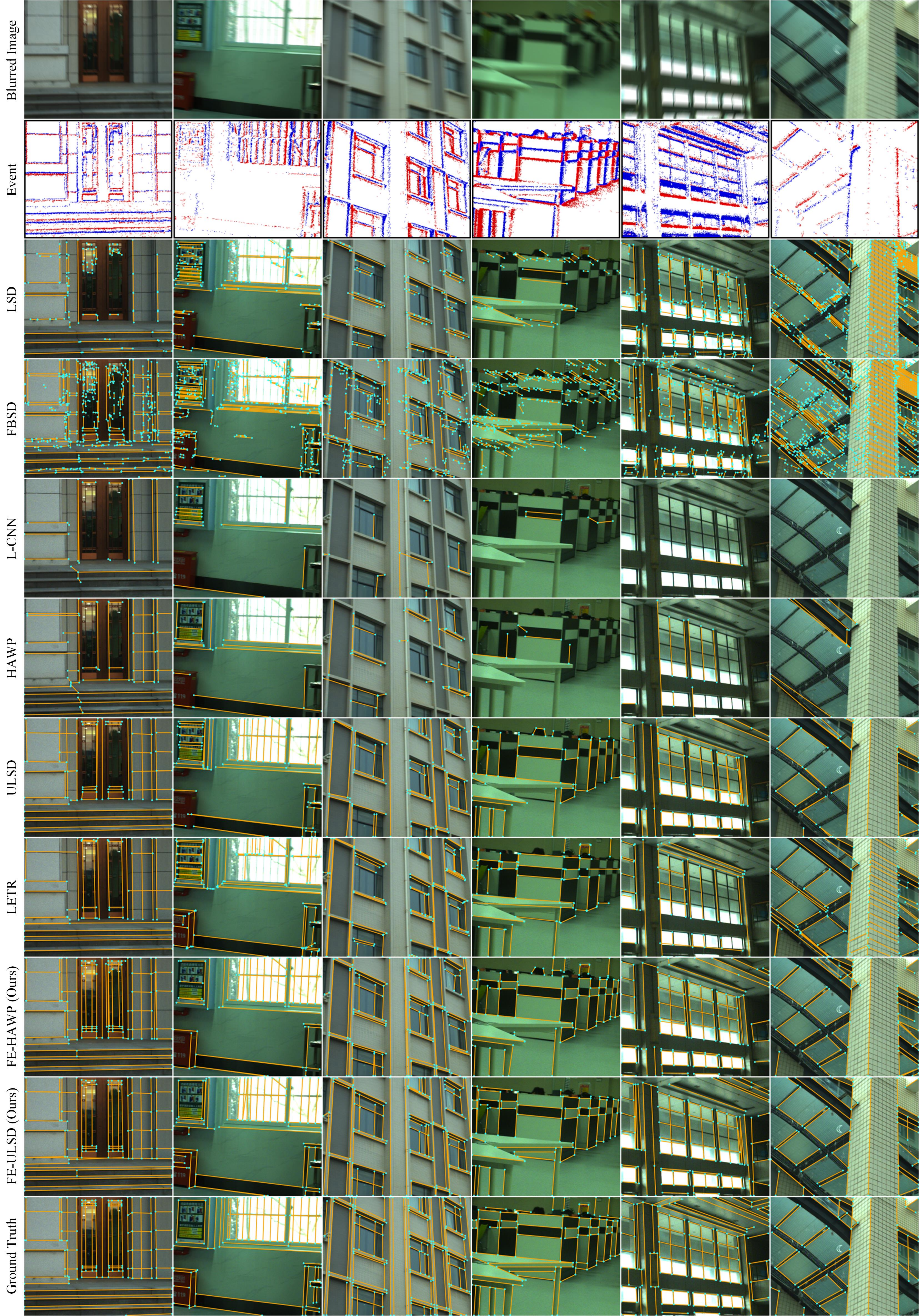}
	\caption{Result comparisons on real \whu dataset (Lines are plotted on clear images but detected on the motion-blurred images and events)}
	\label{fig:whucomparison}
\end{figure*}

Fig. \ref{fig:whucomparison} shows the qualitative results on \whu ~dataset. The conventional LSD and FBSD have consistent performances as on the synthetic FE-Wireframe dataset. The officially trained models, HAWP and LETR, still have large missing detection but with higher accuracy than the same model on the synthetic FE-Wireframe dataset. This is because of the significant data differences between synthetic and real-world datasets. Then retraining on the concatenation of images and events greatly improves the performance and yields more detected line segments. Similar to the results on FE-Wireframe, the proposed FE-HAWP and FE-ULSD detect the most line segments with the highest similarity to the ground truth labels, reflecting the effectiveness of the proposed method.

\subsection{Discussions}
\begin{figure}[htbp]
	\begin{minipage}[t]{0.49\linewidth}
		\centering
		\includegraphics[width=\columnwidth]{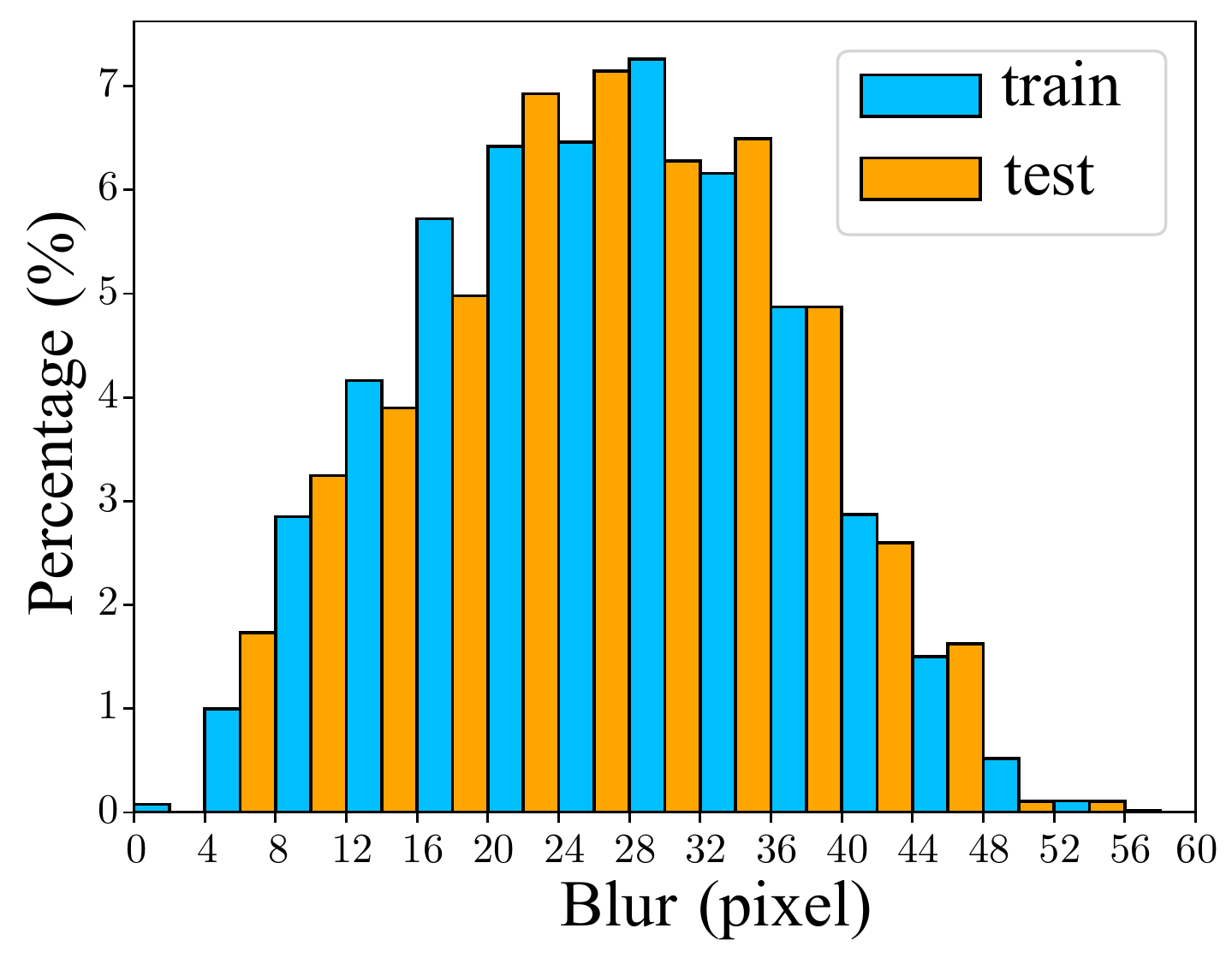} 
	\end{minipage}
	\begin{minipage}[t]{0.49\linewidth} 
		\centering
		\includegraphics[width=\columnwidth]{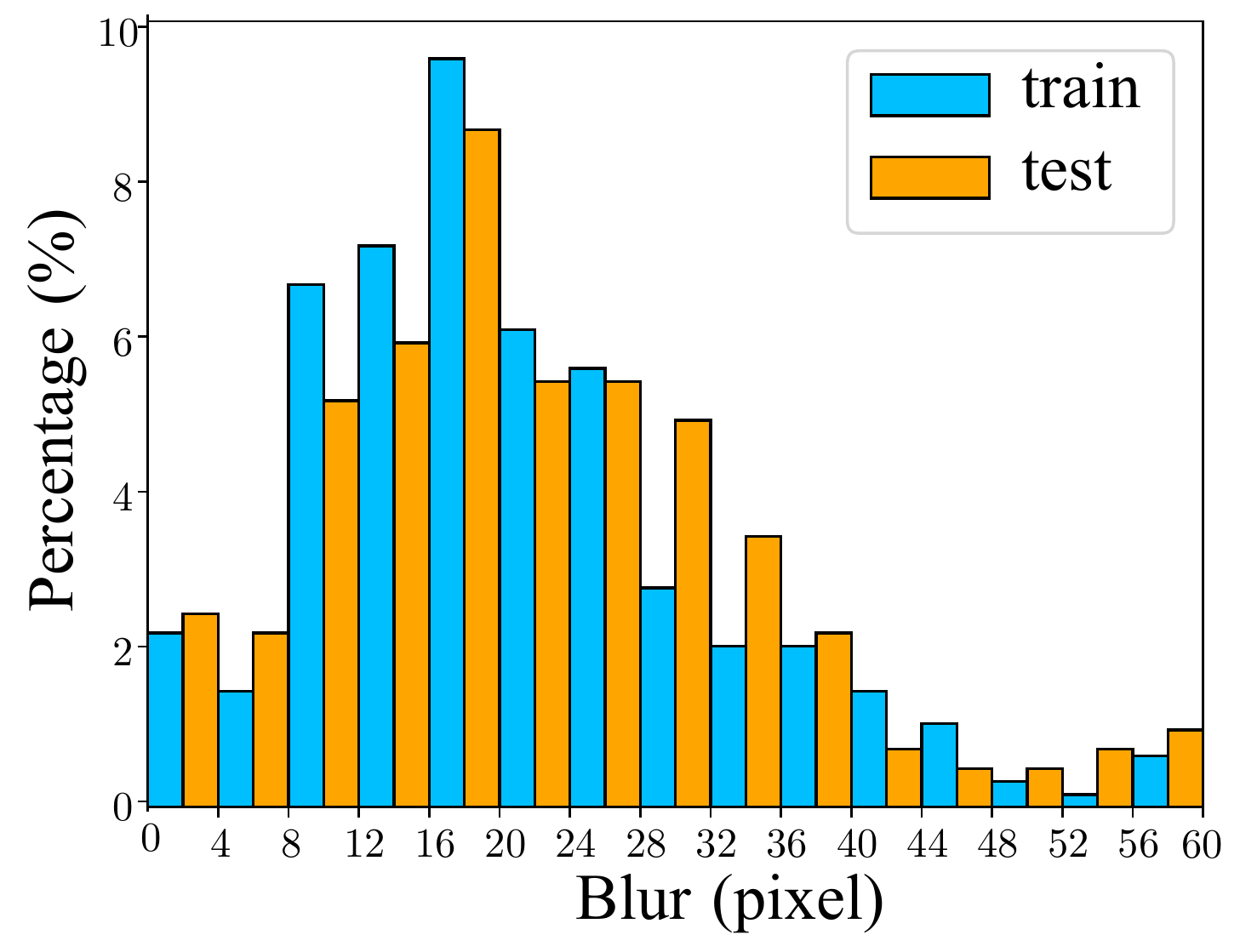}
	\end{minipage}
	\begin{minipage}[t]{0.50\linewidth} 
		\centering
		\includegraphics[width=\columnwidth]{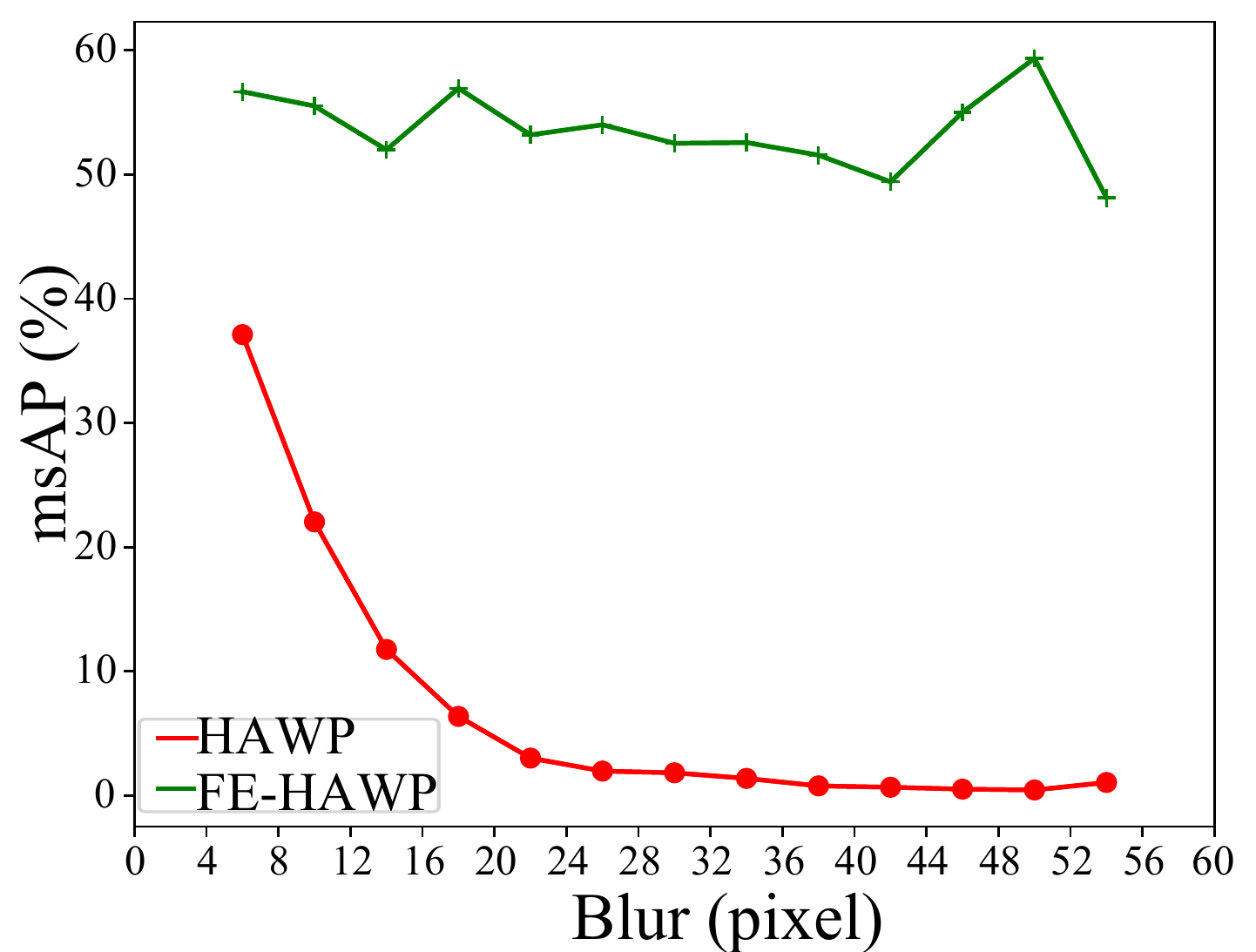}
	\end{minipage}
	\begin{minipage}[t]{0.50\linewidth} 
		\centering
		\includegraphics[width=\columnwidth]{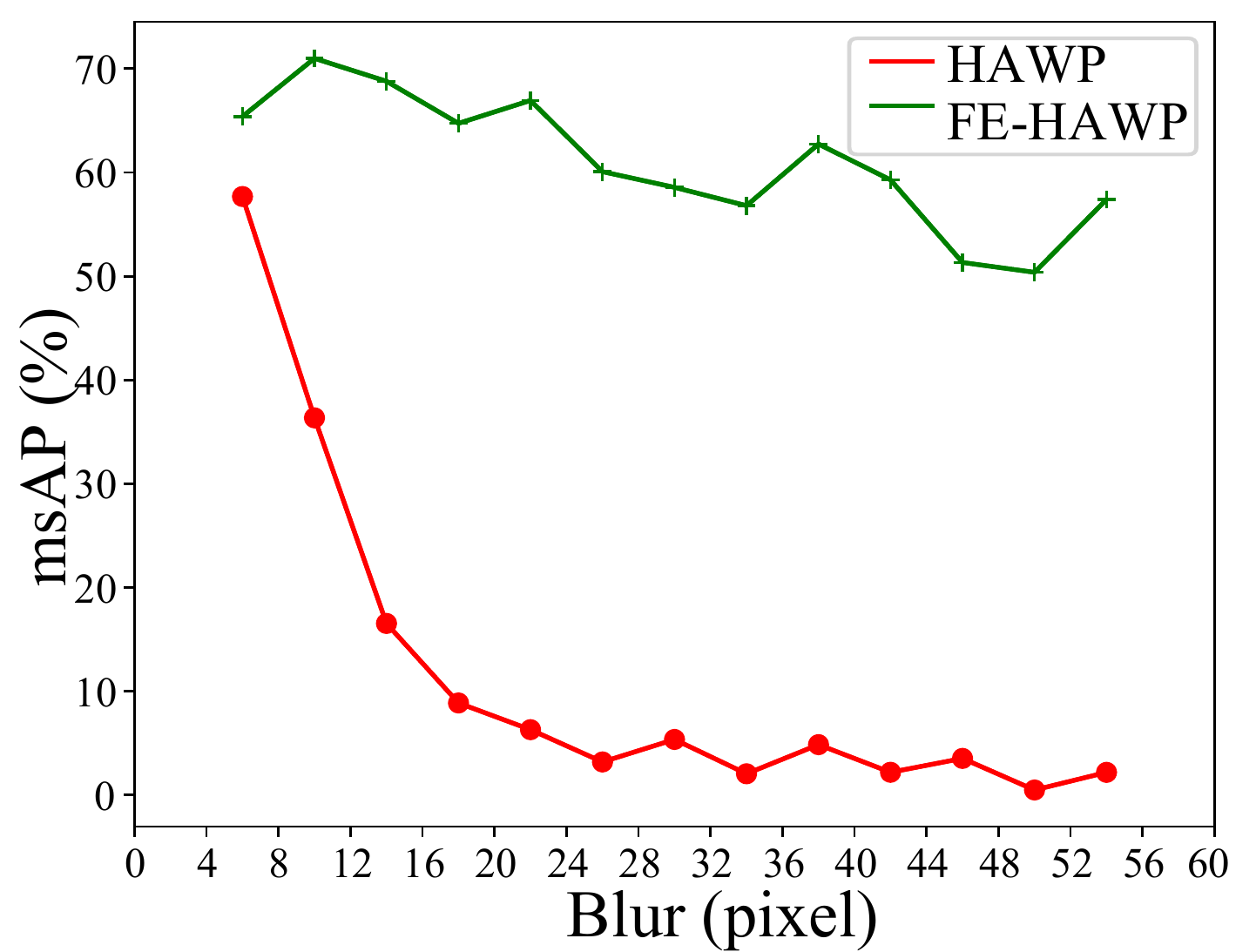}
	\end{minipage}
	\caption{Motion blur distributions and the line detection results upon the displacements. Top-left: Motion blur distributions on FE-Wireframe; Top-right: Motion blur distributions on \whu; Bottom-left: msAP distributions of HAWP and FE-HAWP on FE-Wireframe; Bottom-right: msAP distributions of HAWP and FE-HAWP on \whu.}
	\label{fig:blurs}
\end{figure}

By fusing image and event data, we effectively improve the performance of line segment detection on motion-blurred images. However, it is still hard to determine the performance of the fusion over diverse camera motions. Therefore, we analyze the motion blur degree by counting the average displacement of junctions. The distributions of displacement and the corresponding line detection performance are plotted in Fig.~\ref{fig:blurs}, where FE-HAWP and HAWP are evaluated. The x-axis represents the blurriness, which is computed by averaging junction flow from the start image to the end image for each data. It can be observed that the blurriness is between 0-60 pixels, and the blur distributions of the synthetic FE-Wireframe and the real \whu are slightly different. The msAP distributions of official HAWP indicate that larger blurs will result in lower line detection accuracy. Then when using the proposed FE-HAWP on the fusion of images and events, the line detection accuracy is much higher and more robust to the different blurriness. 
\section{Conclusion}
\label{sec::conclusion}

In this paper, we have presented a line segment detection method using events to address its performance degradation in motion-blurred images. The proposed frame-event feature fusion backbone fully exploits the complementary information between images and event data with shallow and multi-scale decoder fusion. Therefore, the structural edge information can be well extracted in the feature map regardless of slow or fast camera motions. The state-of-the-art line segment detectors, HAWP and ULSD, are then employed to conduct the end-to-end line segment detection. To train our model and contribute to the community, two frame-event line segment detection datasets, FE-Wireframe and \whu, are constructed with motion-blurred images and spatially-temporally-aligned event data. Extensive component configuration analyses validate the reasonability of our design, and comprehensive comparisons to the state-of-the-arts demonstrate the effectiveness of the proposed method. In the future, we will further explore the line endpoints flow in event streams to facilitate line segment detection and improve efficiency.

\ifCLASSOPTIONcaptionsoff
  \newpage
\fi

\bibliographystyle{IEEEtran}
\bibliography{tpami-name,main}

%



\end{document}